\documentclass{ifacconf}

\usepackage{graphicx}      
\usepackage{natbib}      
\usepackage{makecell}
\usepackage{amsmath}
\usepackage{bm}
\usepackage{algorithm}
\usepackage{algpseudocode}
\usepackage{amsfonts}
\usepackage[dvipsnames]{xcolor}
\usepackage{url}
\usepackage{multirow}
\usepackage{mathtools}
\usepackage{footnote}

\usepackage{textpos} 

\makesavenoteenv{tabular}
\mathtoolsset{showonlyrefs}
\begin{document}
\begin{frontmatter}

\begin{textblock*}{3cm}(-4cm,-2.2cm)
   \fbox{\footnotesize $\copyright$ 2020 F. Zocco and S. McLoone. \emph{This work has been accepted to IFAC for publication under a Creative Commons Licence CC-BY-NC-ND}}
\end{textblock*}

\title{An Adaptive Memory Multi-Batch L-BFGS Algorithm for Neural Network Training} 

\thanks[footnoteinfo]{The first author gratefully acknowledges the financial support provided by Irish Manufacturing Research (IMR) for this research.}

\author{Federico Zocco\thanksref{footnoteinfo} } 
\author{\,\,\,\,\,\,Se\'an McLoone} 

\address{Centre for Intelligent Autonomous Manufacturing Systems, Queen's University Belfast, Northern Ireland, UK
(e-mail: \{fzocco01, s.mcloone\}@qub.ac.uk)}

\begin{abstract}                
Motivated by the potential for parallel implementation of batch-based algorithms and the accelerated convergence achievable with approximated second order information a limited memory version of the BFGS algorithm has been receiving increasing attention in recent years for large neural network training problems. As the shape of the cost function is generally not quadratic and only becomes  approximately quadratic in the vicinity of a minimum, the use of second order information by L-BFGS can be unreliable during the initial phase of training, i.e. when far from a minimum. Therefore, to control the influence of second order information as training progresses, we propose a multi-batch L-BFGS algorithm, namely MB-AM, that gradually increases its trust in the curvature information by implementing a progressive storage and use of curvature data through a development-based increase (\emph{dev-increase}) scheme. Using six discriminative modelling benchmark problems we show empirically that MB-AM has slightly faster convergence and, on average, achieves better solutions than the standard multi-batch L-BFGS algorithm when training MLP and CNN models. 
\end{abstract}    

\begin{keyword}
Deep learning, L-BFGS, variable memory, quasi-Newton methods, neural networks
\end{keyword}

\end{frontmatter}
\section{Introduction}
In the last twenty years significant advances have been made towards making artificial neural networks able to compete with their biological counterparts (\cite{dodge2017study}). A factor that is critical to the performance of artificial neural networks is the network training algorithm, which determines the sequence of computations to be performed in order for the network to learn the underlying relationships in the dataset of interest. Many training algorithms have been proposed over the years to achieve this goal, typically trading off computational complexity with rate-of-convergence and/or quality of solutions obtained (\cite{mcloone1998hybrid,goodfellow2016deep,ruder2016overview}). In general, training algorithms fall into three categories: first order methods, which calculate the loss function and its derivative at each iteration, e.g. stochastic gradient descent (SGD) and Adam (\cite{kingma2014adam}), second order methods, which also calculate second derivative information, and quasi-Newton methods, which evaluate an approximation of the second derivative instead of computing it directly, e.g. the limited-memory Broyden-Fletcher-Goldfarb-Shanno (L-BFGS) (\cite{liu1989limited}). Whenever the amount of data to be processed is large, state-of-the-art performance is usually achieved with well-tuned first order methods thanks to regularization techniques such as batch normalization and dropout (\cite{bollapragada2018progressive}). Regularization and stable approximated curvature updates in L-BFGS methods are currently an active area of research due to the accelerated convergence achievable with curvature information and the ability to exploit parallelism with large batch sizes to achieve efficient algorithm implementations (\cite{berahas2016multi, yousefian2017stochastic}). \cite{agarwal2014reliable} proposed a hybrid approach where training is initially performed with a fast and regularized first order method and then switched to a full batch method to exploit parallelism. The L-BFGS method we propose can be seen as a hybrid approach where the batch size is kept fixed as in \cite{berahas2016multi}, but the use of approximated second order information is gradually increased during training thereby enhancing convergence. Initially, when little second order information is used, the method is essentially first order and becomes increasingly second order in nature as more and more curvature information is incorporated into the weight updates.      

\textbf{Our contributions}: Motivated by recent interest in progressive training strategies (\cite{smith2017don,bollapragada2018progressive}), we propose an approach based on \emph{progressive curvature trust} for the multi-batch L-BFGS algorithm of \cite{berahas2016multi} by adding two algorithmic ingredients: a progressive storage (and use) of curvature information, and periodic resetting. Based on the development-based decay (\emph{dev-decay}) scheme for first order methods (\cite{wilson2017marginal}), we propose a \emph{dev-increase} scheme to control the progression of the memory size for curvature information and its resetting. For comparison purposes, we develop three variants of the multi-batch L-BFGS and experimentally compare them with the multi-batch L-BFGS and Adam algorithms for training multilayer perceptrons (MLPs) and convolutional neural networks (CNNs) on six case study datasets.


The paper is organized as follows: Section \ref{sec:BasicAlg} describes the multi-batch L-BFGS algorithm of \cite{berahas2016multi} as this is the reference algorithm to which we add two algorithmic ingredients; Section \ref{sec:Main} explains the concept of \emph{progressive curvature trust} and the proposed algorithm; Section \ref{sec:Experiments} describes the case studies and experiments conducted to evaluate the proposed algorithm and discusses the results obtained. Finally, Section \ref{sec:Concl} gives the conclusions. 

Hereinafter, matrices and  vectors are indicated with bold capital and bold lowercase letters, while lowercase italic font denotes scalars. The sets are indicated by italic capital letters.

\section{Multi-Batch L-BFGS}\label{sec:BasicAlg} 
Let us consider a classification task defined by the set 
\begin{equation}
\mathcal{B} = \{\bm{x}^{(i)}, l^{(i)}\}_{i=1,2,\dots,|\mathcal{B}|}, 
\end{equation}
where $\bm{x}^{(i)} \in \mathbb{R}^{n}$ is an input sample (e.g. an image) of size $n$ and $l^{(i)} \in \mathbb{N}$ is an integer representing its class (i.e. label). A neural network relates the input to the output label via a mapping $p(\bm{\theta}, \bm{x}^{(i)}): \mathbb{R}^{n} \mapsto \mathbb{N}$, with network parameters $\bm{\theta} \in \mathbb{R}^{d}$ optimally adapted to $\mathcal{B}$ using a training algorithm. Full-batch algorithms learn $\bm{\theta}$ by considering the complete dataset $\mathcal{B}$, therefore they optimize a deterministic function $C(\mathcal{B})$, that is
\begin{equation}
\min_{\bm{\theta}} \frac{1}{|\mathcal{B}|}\sum_{i \in \mathcal{B}}{c(\bm{x}^{(i)}, l^{(i)}; p(\bm{\theta}, \bm{x}^{(i)}))} = \min_{\bm{\theta}} C(\mathcal{B}),
\end{equation}
where $c(\cdot)$ is a cost (i.e. loss) function chosen to measure the distance between the model prediction $\hat{l}^{(i)} = p(\bm{\theta}, \bm{x}^{(i)})$ and the expected output $l^{(i)}$ when given the sample $\bm{x}^{(i)}$ as input.  
In contrast, in multi-batch mode a random subset $\mathcal{S}_k \subset \mathcal{B}$ is used at each iteration and the optimal $\bm{\theta}$ is estimated by iteratively minimizing a stochastic cost function $C(\mathcal{S}_k)$, that is
\begin{equation}
\min_{\bm{\theta}_k} \frac{1}{|\mathcal{S}_k|}\sum_{i \in \mathcal{S}_k}{c(\bm{x}^{(i)}, l^{(i)}; p(\bm{\theta}, \bm{x}^{(i)}))} = \min_{\bm{\theta}_k} C(\mathcal{S}_k),
\label{eq:Ck}
\end{equation}
where $k$ is the iteration counter. The L-BFGS algorithm updates the parameters with the rule
\begin{equation}
\bm{\theta}_{k+1} = \bm{\theta}_k - \eta \bm{H}_k^{-1} \bm{g}_k^{\mathcal{S}_k}
\label{eq:wUpdate}
\end{equation}
where $\eta$ is the learning rate (i.e. step length), $\bm{g}_k^{\mathcal{S}_k}$ is the gradient defined as
\begin{equation}
\bm{g}_k^{\mathcal{S}_k} = \frac{\partial C(\mathcal{S}_k)}{\partial \bm{\theta}_k}
\label{eq:gradDef}
\end{equation}
and $\bm{H}_k^{-1}$ is the inverse Hessian matrix approximation updated according to (\cite{nocedal2006numerical})
\begin{align}
\bm{H}_{k+1}^{-1} &= \bm{V}_k^\top \bm{H}^{-1}_k \bm{V}_k + \rho_k \bm{s}_k \bm{s}_k^\top \\
\rho_k &= (\bm{t}_k^\top \bm{s}_k)^{-1} \\
\bm{V}_k &= \bm{I} - \rho_k\bm{t}_k \bm{s}_k^\top \\
\bm{s}_k &= \bm\theta_{k+1} - \bm{\theta}_k  \label{eq:updateOfs}\\
\bm{t}_{k} &= \bm{g}^{\mathcal{S}_k}_{k+1} - \bm{g}^{\mathcal{S}_k}_k. \label{eq:updateOft}
\end{align}
The storage of $\bm{H}_k^{-1}$ requires $\mathcal{O}(d^2)$ memory, which is prohibitive with large neural networks, therefore in L-BFGS a two-loop recursion is used to directly compute the product $\bm{H}_k^{-1} \bm{g}_k^{\mathcal{S}_k}$ such that only a predefined number of curvature pairs ($\bm{s}_k$, $\bm{t}_{k}$) need to be stored. The maximum number of stored pairs, $m$, is usually fixed between 3 and 20 (\cite{nocedal2006numerical}). Once the limit $m$ is reached the oldest pairs are replaced by the newest ones. Given the stochastic nature of the evaluation of the gradient using \eqref{eq:gradDef}, to achieve a stable computation of the curvature $\bm{t}_k$ (see \eqref{eq:updateOft}) \cite{berahas2016multi} proposed having overlapping consecutive batches, i.e. $\mathcal{O}_{k} = \mathcal{S}_{k-1} \cap \mathcal{S}_{k} \neq \emptyset$. This overlap is defined with the hyperparameter $o = \frac{|\mathcal{O}_k|}{|\mathcal{S}_k|}$, which is usually chosen in the range $0 < o < 0.5$.

\section{Adaptive Memory Multi-Batch L-BFGS}\label{sec:Main}
\subsection{Local Approximation of the Cost Function} 
An important feature of gradient-based optimization algorithms is the computation of the search direction used to find an updated parameter vector $\bm{\theta}_{k+1}$. Newton and quasi-Newton methods such as L-BFGS define the direction approximating the exact cost function $C$ with a quadratic model, obtained by truncating the cost function Taylor series expansion at the third term, that is:
\begin{equation}
\begin{split}
C(\bm{\theta}_k + \bm{y}_k) \approx \bar{C}_k = \hspace{0.4in}\\
= C(\bm{\theta}_k) + \bm{g}_k^\top(\bm{\theta}_k) \bm{y}_k + \frac{1}{2}{\bm{y}_k^\top \bm{H}_k(\bm{\theta}_k) \bm{y}_k}.
\end{split}
\end{equation}
Then, differentiating with respect to $\bm{y}_k$ and setting it equal to zero we find the direction leading to the minimum of $\bar{C}$
\begin{equation}
\bm{y}_k = - \bm{H}_k^{-1} \bm{g}_k, 
\end{equation} 
which motivates the weight update rule in \eqref{eq:wUpdate}. However, if around the point $\bm{\theta}_k + \bm{y}_k$ the function is significantly different from a quadratic-like model, the use of the curvature information could be detrimental to stable and reliable parameter updates. Moreover the curvature evaluation is computationally demanding as it requires the computation of $d(d+1)/2$ derivatives, thus it is only worth using if it actually improves the algorithm convergence.

Now we observe that as we approach a minimum point of a nonlinear multi-modal function the quadratic approximation becomes increasingly valid, and this occurs at the more advanced stages of training, i.e. 
\begin{equation}
\lim_{k\to\infty} C-\bar{C}_k = 0.
\end{equation}
In contrast, at the beginning of the training process it is difficult to say how well a quadratic model will approximate the true cost function.

Therefore, we propose an algorithm that implements a progressive use of curvature information: initially only a few curvature pairs (e.g. one or two) are stored and used, with the older pairs discarded to reduce the computational cost; then, when the algorithm gets closer to the minimum, we increase the storage and use of second order information.

\subsection{The Algorithm} 
The pseudo-code of the proposed method is presented in Algorithm \ref{pseudocode}, which is the result of the combination of three building blocks: 

\textbf{Block 1:} The basic structure is the multi-batch L-BFGS of \cite{berahas2016multi} described in Section \ref{sec:BasicAlg} which performs stable curvature updates by overlapping consecutive batches. A PyTorch implementation is available on GitHub\footnotemark. 

\textbf{Block 2:} To regulate the use of approximated second order information during training the memory $m_k$ used to store the curvature pairs is increased as the number of iterations increases, in a fashion similar to the progressive batching approach proposed in \cite{bollapragada2018progressive}. The rule defining when to increase $m_k$ is based on the development of the validation loss in a similar fashion to the \emph{dev-decay} scheme of \cite{wilson2017marginal} applied to the step length. Adapted to implement a progressive curvature trust, it becomes a \emph{dev-increase} scheme. It is executed between Step 11 and Step 18, with $\alpha$ denoting the memory scaling factor. The macro-condition \eqref{macroCond} is typically satisfied when the validation loss is approaching a local minimum. The parameter $m_{val}$ defines how many previous validation losses are stored and, if increased, makes the satisfaction of $Q$ more difficult, thus delaying the increase of $m_k$. The lower and upper limits of $m_k$ are $m_{0}$ and $m_{max}$, respectively, i.e. $m_0 \leq m_k \leq m_{max}$. Note that the extra memory required to store the validation losses is negligible with large datasets/models because $v_k$ is a scalar and $m_{val}$ is a relatively small number (e.g. $m_{val} \leq 10$). 
\begin{equation}
Q: \{\Delta_{k-m_{val}+j} > \Delta_{k-m_{val}+j+1}\}_{j=0,1,\dots,m_{val}-2} 
\label{macroCond}
\end{equation}
with
\begin{equation}
\Delta_k = v_{k-1} - v_k.
\label{eq:Delta}
\end{equation}

\textbf{Block 3:} To further mitigate the impact of using unrepresentative curvature information, the resetting of the memory proposed in \cite{mcloone1999variable} is implemented. Adapted to implement a progressive curvature trust, the resetting is applied just for smaller $k$ according to the development of $m_k$, which in turn follows the validation loss development. This third building block is implemented between Step 19 and Step 21, where $q_k$ is the current number of stored curvature pairs and $m_{reset}$ is the memory size below which resetting is applied.

\begin{algorithm}
\centering
\caption{Adaptive Memory Multi-Batch L-BFGS}
\label{pseudocode}
\begin{algorithmic}[1]
\State \textbf{Input:} $\bm{\theta}_0$ (initial iterate), $\mathcal{B} = \{(\bm{x}^{(i)}, l^{(i)})$, for $i \in \mathcal{B}$\} (training set), $r$ (i.e. $|\mathcal{S}_k|$), $o$ (overlap ratio), $\eta$, $\alpha$, $m_0$, $m_{max}$, $m_{val}$, $m_{reset}$ 
\State \textbf{Initialisation:} $k \leftarrow 0$, $q_{k} \leftarrow 0$, $m_k \leftarrow m_0$
\State Randomly select a batch $\mathcal{S}_0 \subset \mathcal{B}$
\State \textbf{Repeat} until convergence
\State Compute $C_k$ 
\State Compute $\bm{g}^{\mathcal{S}_k}_k$
\State Compute $\bm{H}_k \bm{g}^{\mathcal{S}_k}_k$ via L-BFGS two-loop recursion (\cite{nocedal2006numerical})
\State Update the parameters as in \eqref{eq:wUpdate}
\State Select a new batch $\mathcal{S}_{k+1} \subset \mathcal{B}$ with overlap ratio $o$
\State Compute the new curvature pair ($\bm{s}_k$, $\bm{t}_k$) as in \eqref{eq:updateOfs}, \eqref{eq:updateOft}
\State Compute the new validation loss $v_k$
\If{number of stored $v_k$ == $m_{val}$}
	\State Remove the oldest $v_k$
\EndIf
\State Store the newest validation loss $v_k$ 
\If{$m_k < m_{max} \quad \& \quad Q$}
	\State $m_k \leftarrow \alpha m_k$ 
\EndIf
\If{$q_{k} == m_k$}
	\If{$m_k \leq m_{reset}$}
	\State Delete all stored pairs and $q_{k} \leftarrow 0$ 
\Else 
\State Remove the oldest pair and $q_{k} \leftarrow q_{k}-1$
\EndIf
\EndIf 
\State Store the newest pair and $q_{k} \leftarrow q_{k}+1$
\State $k \leftarrow k+1$
\end{algorithmic}
\end{algorithm}

\section{Experiments}\label{sec:Experiments}
\subsection{Experimental Setup}
In this section two implementations based on Algorithm \ref{pseudocode} are considered: multi-batch L-BFGS with adaptive memory and without resetting (MB-AM), i.e. $m_{reset} = 0$, and multi-batch L-BFGS with both adaptive memory and resetting (MB-AMR), i.e. $m_{reset} > 0$. In addition, a multi-batch L-BFGS with constant memory and periodic resetting, as proposed in \cite{mcloone1999variable}, is considered. This variant, denoted as MB-R, performs resetting whenever the maximum memory size is reached. For benchmarking purposes, two additional methods are considered, namely, standard multi-batch L-BFGS (MB), and Adam. Therefore in total five methods are compared. 

The six datasets and six experiments considered are detailed in Table \ref{tab:datasetsDescription} and Table \ref{tab:ExpDescription}, respectively. The MLP is a single-hidden layer fully connected feedforward neural network with $h$ neurons in the hidden layer. The CNN has two 2D convolutional layers followed by two fully connected layers. Each convolutional layer is followed by a ReLU and a 2D max pooling layer. The function $C(\cdot)$ in \eqref{eq:Ck} minimized during training is the cross-entropy loss function  (\cite{bishop2006pattern}), with $c(\cdot)$ defined as
\begin{equation}
c(l^{(i)}, \hat{l}^{(i)}) = - \{l^{(i)} \ln(\hat{l}^{(i)}) + (1 - l^{(i)}) \ln(1 - \hat{l}^{(i)})\}. 
\end{equation}
The algorithms were written in PyTorch. The data augmentation of the TRASH dataset was performed using Keras. The experiments with the CNN were executed on an nVidia P40-4Q virtual GPU, while the experiments with the MLP were executed on a local desktop with an Intel i7-6700 CPU and 16 GB of RAM. The code is available on GitHub\footnotemark. 

\begin{table*}
\begin{center}
\caption{Overview of the case study datasets.}
\label{tab:datasetsDescription}
\begin{tabular}{lcccl} 
Dataset & \# samples (train; test) & \# features & \# classes & Source \\
\hline
CANCER & (484; 85) & 30 & 2 & \cite{street1993nuclear}\footnotemark\\ 
ETCH & (1865; 329)  & 2046 & 3 & \cite{puggini2015extreme}\\ 
WAFERS & (6164; 1000)  & 152 & 2 & \cite{olszewski2001generalized}\footnotemark\\ 
MNIST-R\footnotemark & (60000; 10000)  & 32 $\times$ 32 $\times$ 1 & 10 & \cite{lecun1998gradient}\footnotemark \\ 
CIFAR10 & (50000; 10000) & 32 $\times$ 32 $\times$ 3 & 10 & \cite{krizhevsky2009learning}\footnotemark[\value{footnote}] \\ 
TRASH-RA\footnotemark & (45107; 7960) & 32 $\times$ 32 $\times$ 3 & 6 & \cite{yang2016classification}\footnotemark\\ 
\hline
&&&& \\
&&&& \\
\end{tabular}
\end{center}
\end{table*}
\footnotetext[1]{{\tiny \url{https://github.com/hjmshi/PyTorch-LBFGS}}}
\footnotetext[2]{{\tiny \url{https://github.com/fedezocco/AdaMemLBFGS-PyTorch}}}
\footnotetext[3]{{\tiny Downloaded from: \url{https://scikit-learn.org/stable/datasets/}}}
\footnotetext[4]{{\tiny Downloaded from: \url{https://www.cs.ucr.edu/~eamonn/time_series_data/}}}
\footnotetext[5]{{\tiny MNIST-R is a rescaled version of MNIST, which has 28 $\times$ 28 $\times$ 1 features.}}
\footnotetext[6]{{\tiny Downloaded from: \url{https://keras.io/datasets/}}}
\footnotetext[7]{{\tiny TRASH-RA is a rescaled and augmented version of TRASH, which has \hspace{0.5cm} 512 $\times$ 384 $\times$ 3 features and 2527 images in total.}}
\footnotetext[8]{{\tiny Downloaded from: \url{https://github.com/garythung/trashnet}}}
\footnotetext[9]{{\tiny Kernel size = 5$\times$5, kernels per layer = 1, stride = 1, dilatation = 1, padding = 0, pool size = 2$\times$2, pool stride = 2.}}
 
\begin{table*}
\begin{center}
\caption{Description of the design of the six experiments considered, where parameters $h$, $\bar{m}$, $n_{BFGS}$, and $n_{Adam}$ are the number of MLP hidden layer neurons, the maximum memory size for L-BFGS algorithms when $m$ is constant (i.e. MB and MB-R), the number of iterations used for L-BFGS methods, and the number of iterations used for Adam, respectively.}
\label{tab:ExpDescription}
\begin{tabular}{lllc} 
 & Dataset & Model & Hyperparameters\\  
\hline
Experiment 1 & CANCER & MLP & \makecell{$\alpha = 2$, $m_0 = 1$, $m_{max} = 32$, $m_{val} = 5$, $m_{reset} = 8$ $\vert$ $n_{BFGS} = 200$, $o = 0.45$, $\bar{m} = 10$,\\ $r_{BFGS} = 256$, $\eta_{BFGS} = 0.5$ $\vert$ $n_{Adam} = 200$, $r_{Adam} = 64$, $\eta_{Adam} = 0.02$ $\vert$ $h = 35$}\\    
\hline
Experiment 2 & ETCH & MLP & \makecell{$\alpha = 2$, $m_0 = 1$, $m_{max} = 32$, $m_{val} = 5$, $m_{reset} = 8$ $\vert$ $n_{BFGS} = 300$, $o = 0.45$, $\bar{m} = 10$,\\ $r_{BFGS} = 512$, $\eta_{BFGS} = 0.5$ $\vert$ $n_{Adam} = 300$, $r_{Adam} = 64$, $\eta_{Adam} = 0.03$ $\vert$ $h = 320$}\\ 
\hline
Experiment 3 & WAFERS & MLP & \makecell{$\alpha = 2$, $m_0 = 1$, $m_{max} = 32$, $m_{val} = 5$, $m_{reset} = 8$ $\vert$ $n_{BFGS} = 60$, $o = 0.4$, $\bar{m} = 10$,\\ $r_{BFGS} = 512$, $\eta_{BFGS} = 1$ $\vert$ $n_{Adam} = 70$, $r_{Adam} = 64$, $\eta_{Adam} = 0.03$ $\vert$ $h = 5$}\\ 
\hline
Experiment 4 & MNIST-R & CNN\footnotemark & \makecell{$\alpha = 2$, $m_0 = 1$, $m_{max} = 32$, $m_{val} = 5$, $m_{reset} = 8$ $\vert$ $n_{BFGS} = 70$, $o = 0.25$, $\bar{m} = 10$,\\ $r_{BFGS} = 8192$, $\eta_{BFGS} = 1$ $\vert$ $n_{Adam} = 80$, $r_{Adam} = 128$, $\eta_{Adam} = 0.001$}\\ 
\hline
Experiment 5 & CIFAR10 & CNN\footnotemark[\value{footnote}] & \makecell{$\alpha = 2$, $m_0 = 1$, $m_{max} = 32$, $m_{val} = 5$, $m_{reset} = 8$ $\vert$ $n_{BFGS} = 500$, $o = 0.25$, $\bar{m} = 10$,\\ $r_{BFGS} = 8192$, $\eta_{BFGS} = 1$ $\vert$ $n_{Adam} = 1000$, $r_{Adam} = 128$, $\eta_{Adam} = 0.001$}\\ 
\hline
Experiment 6 & TRASH-RA & CNN\footnotemark[\value{footnote}] & \makecell{$\alpha = 2$, $m_0 = 1$, $m_{max} = 32$, $m_{val} = 5$, $m_{reset} = 8$ $\vert$ $n_{BFGS} = 900$, $o = 0.25$, $\bar{m} = 10$,\\ $r_{BFGS} = 8192$, $\eta_{BFGS} = 1$ $\vert$ $n_{Adam} = 1200$, $r_{Adam} = 128$, $\eta_{Adam} = 0.001$}\\ 
\hline
&&& \\
\end{tabular}
\end{center}
\end{table*}

Two metrics are used to assess algorithm performance: the correct classification rate (CCR) and the rank (RNK). CCR is defined as
\begin{equation}
\text{CCR} = \frac{100}{|\mathcal{T}|} \sum_{i \in \mathcal{T}}{z^{(i)}}, \,\,\,z^{(i)} = 
\begin{cases}
      1, & \text{if}\ l^{(i)}=\hat{l}^{(i)} \\
      0, & \text{otherwise}
    \end{cases}  
\end{equation}
where $\mathcal{T}$ is the test set. To evaluate RNK, for each simulation (60 in our experiments) we rank the five candidate methods in descending order such that the one with highest CCR gets RNK = 1, and the one with the lowest CCR gets RNK = 5. Thus, 1 $\leq$ RNK $\leq$ 5.  

\subsection{Results}
\begin{figure}
\begin{center}
\begin{tabular}{c@{\hskip 0.001in}c}
\includegraphics[width=0.245\textwidth]{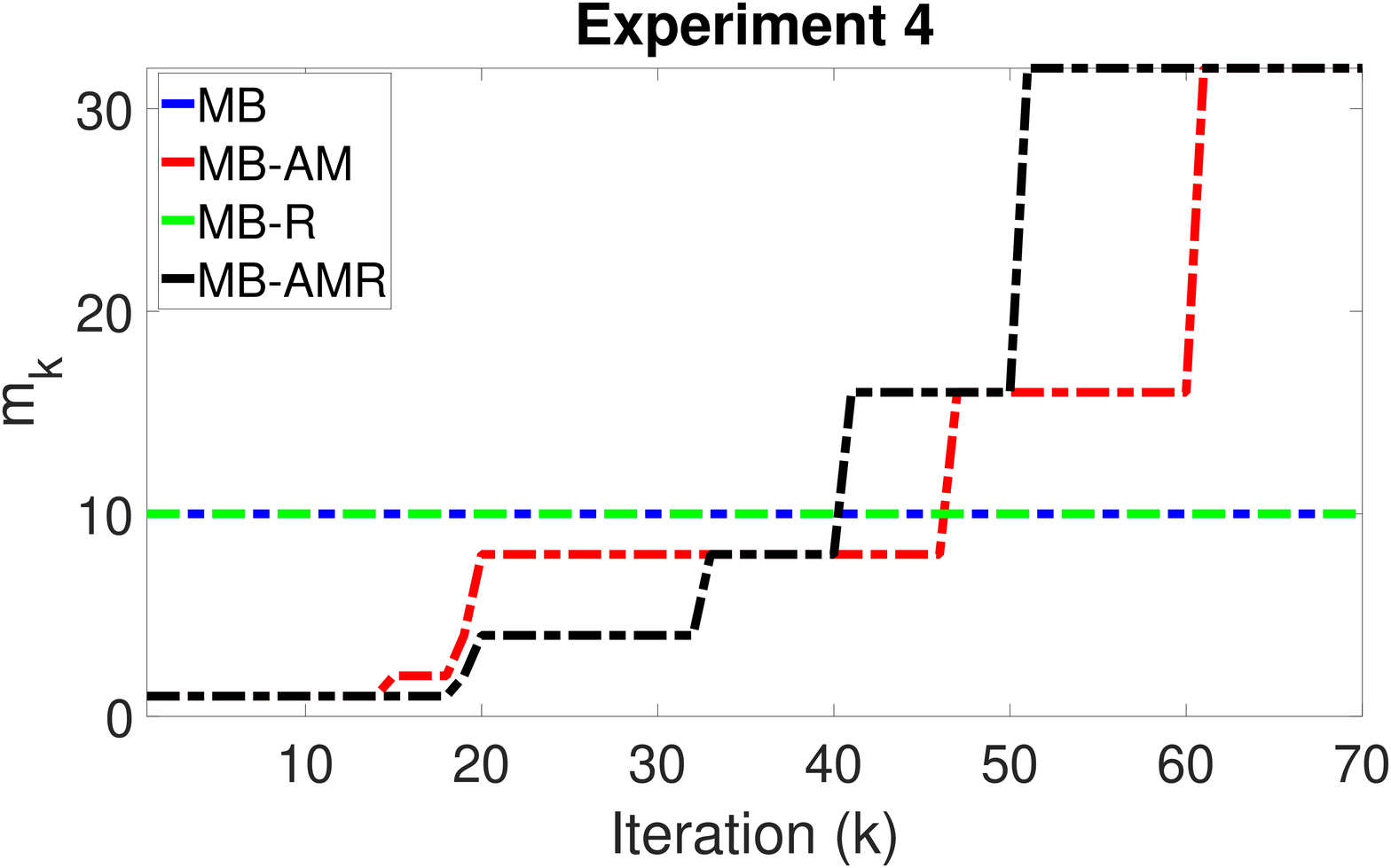} &
\includegraphics[width=0.245\textwidth]{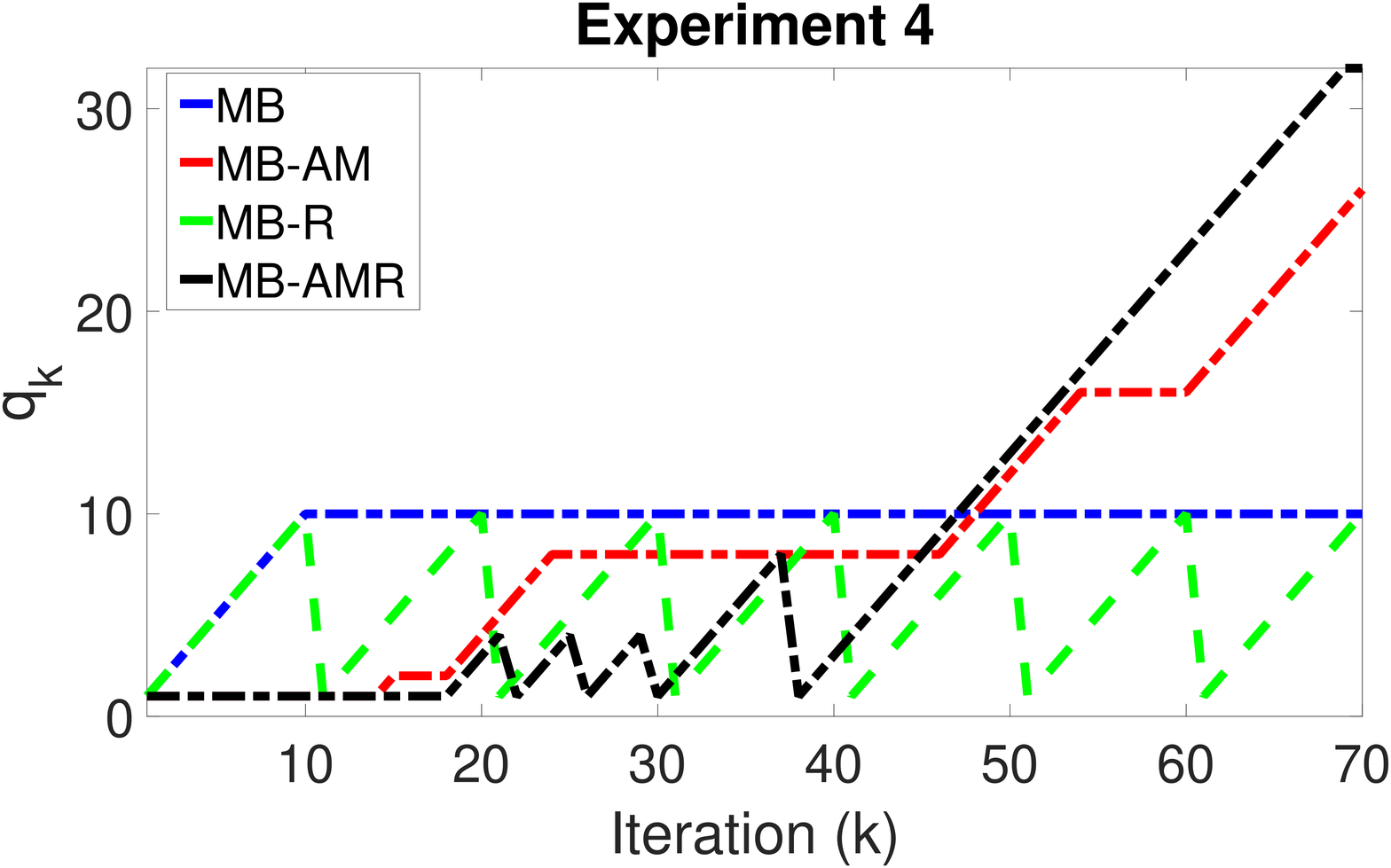} \\
\end{tabular}
\begin{tabular}{c@{\hskip 0.001in}c}
\includegraphics[width=0.245\textwidth]{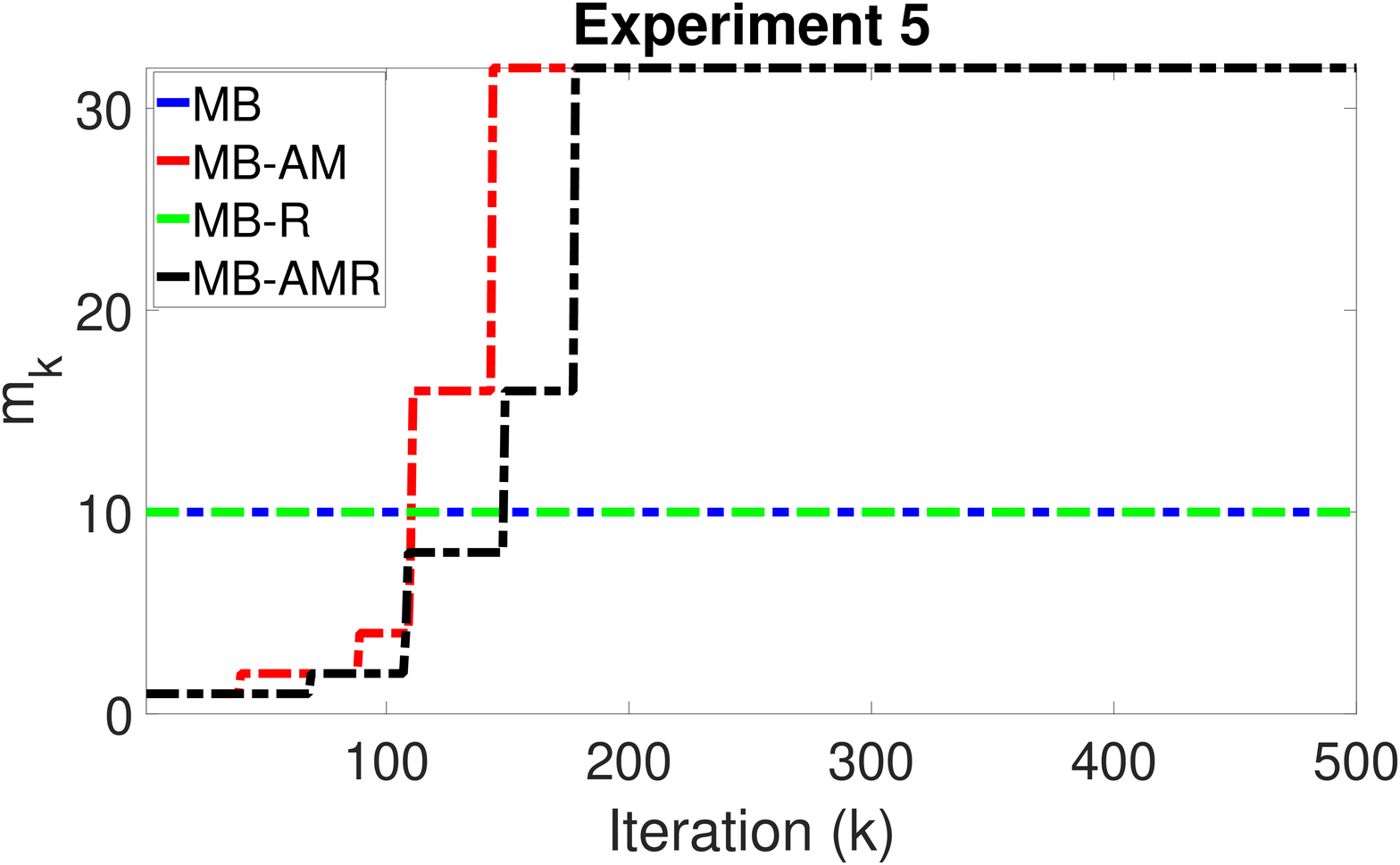} &
\includegraphics[width=0.245\textwidth]{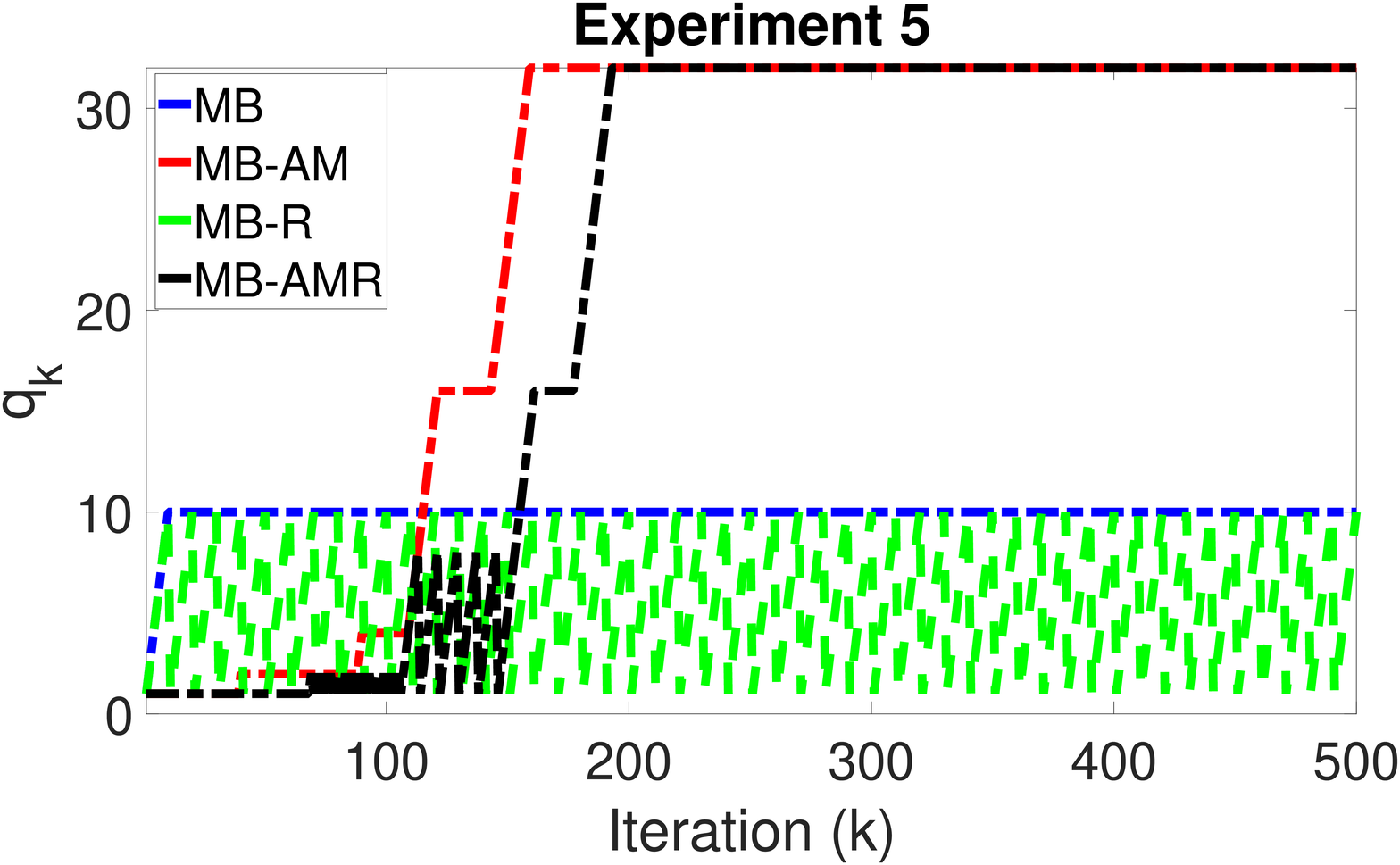} \\
\end{tabular}
\caption{Plot of the memory size $m_k$ (left) and the number of stored curvature pairs $q_k$ (right) as a function of $k$ for selected experiments in Fig. \ref{fig:6x3figure}.}
\label{fig:memorySize}
\end{center}
\end{figure}

\begin{figure*}
\begin{center}
\begin{tabular}{ccc}
\includegraphics[width=0.29\textwidth]{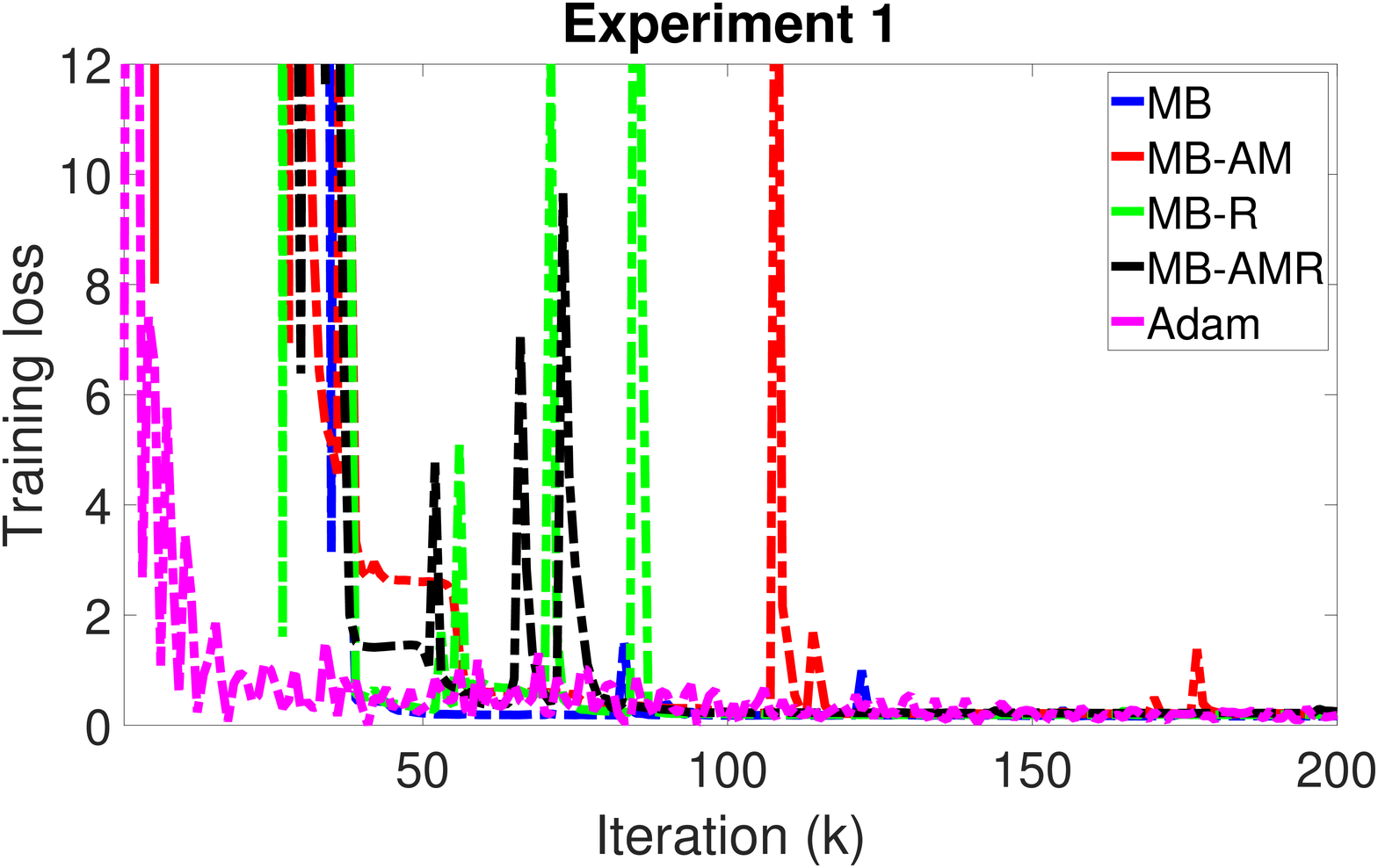} &
\includegraphics[width=0.29\textwidth]{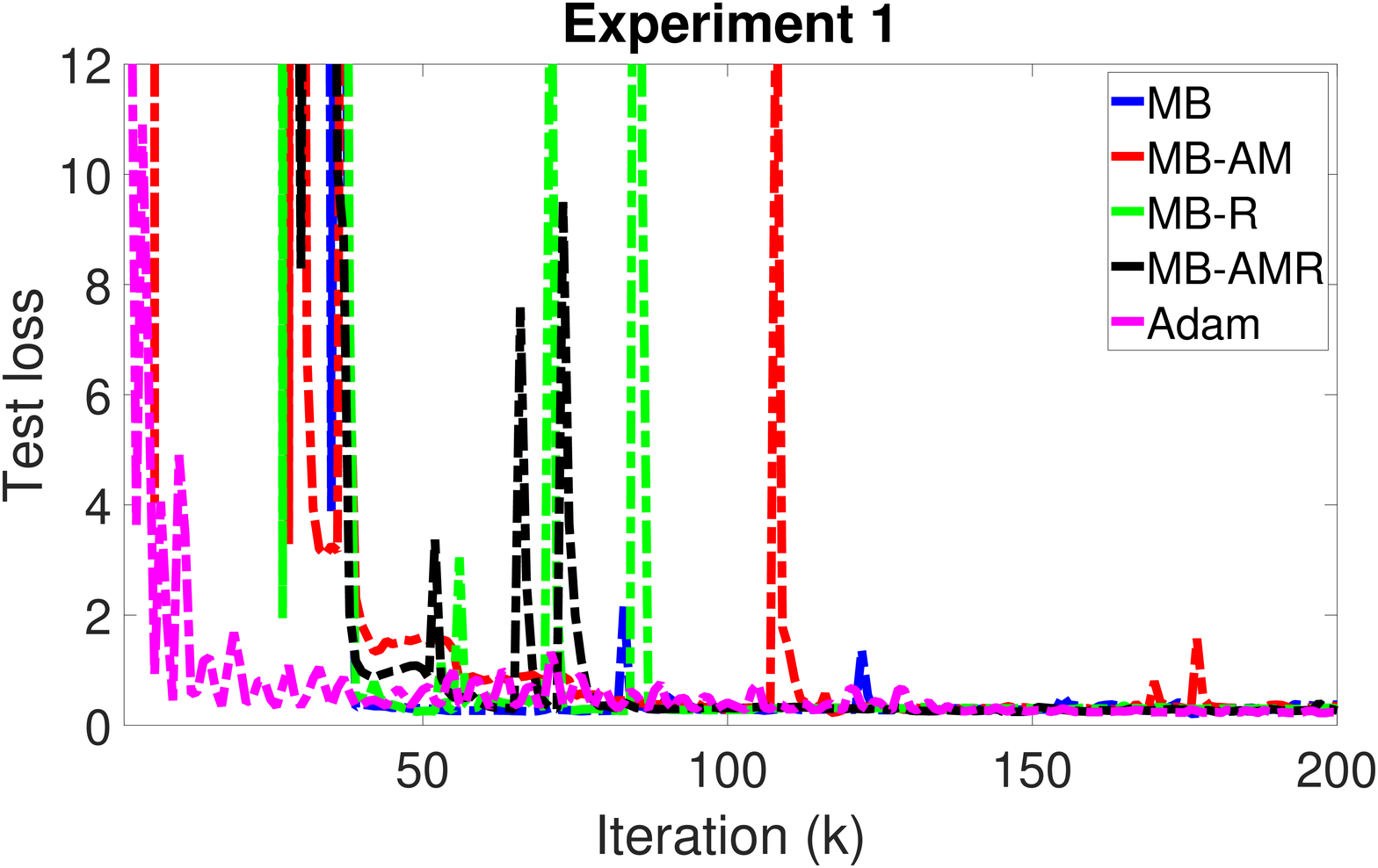} &
\includegraphics[width=0.29\textwidth]{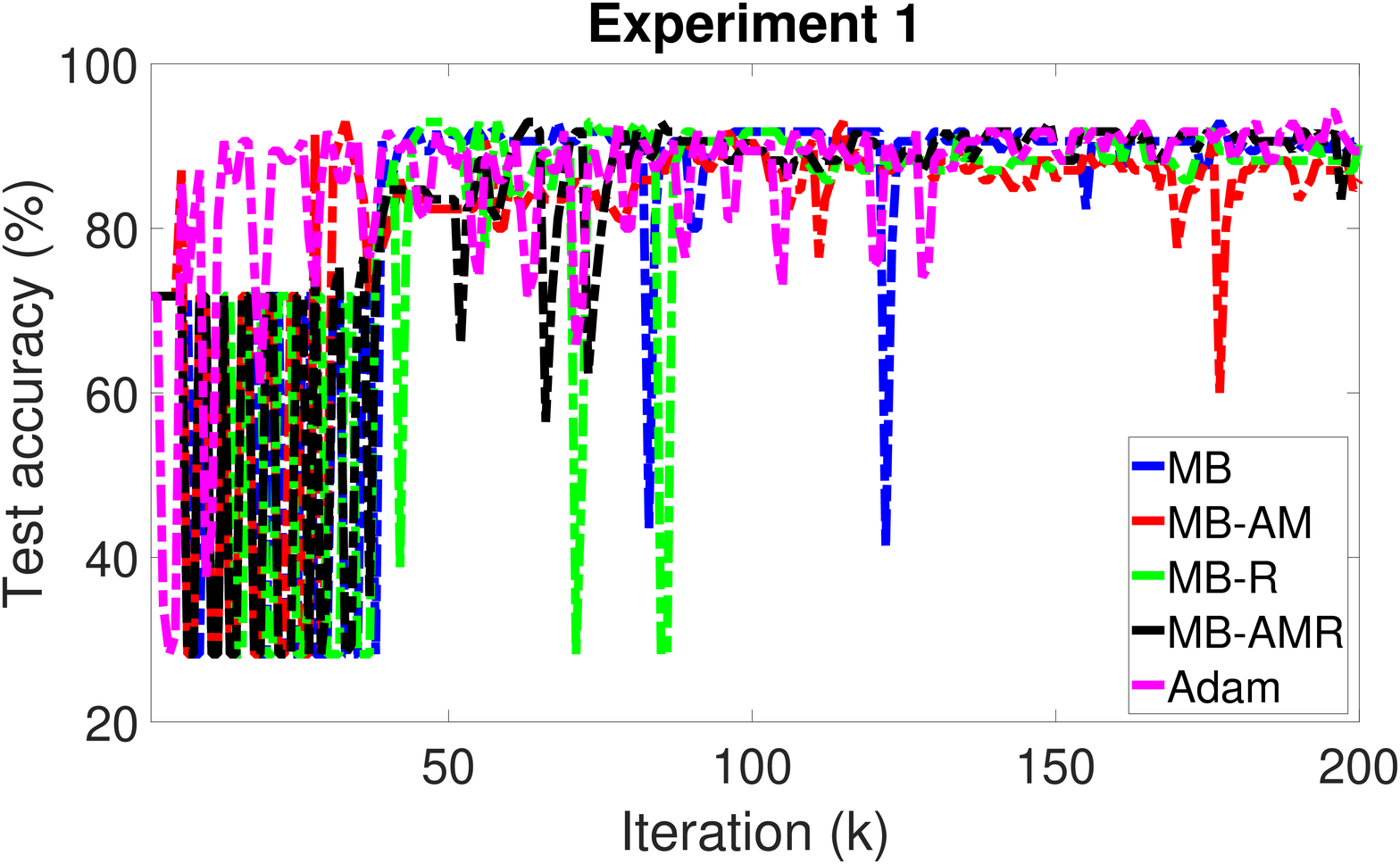} \\
\end{tabular}
\begin{tabular}{ccc}
\includegraphics[width=0.29\textwidth]{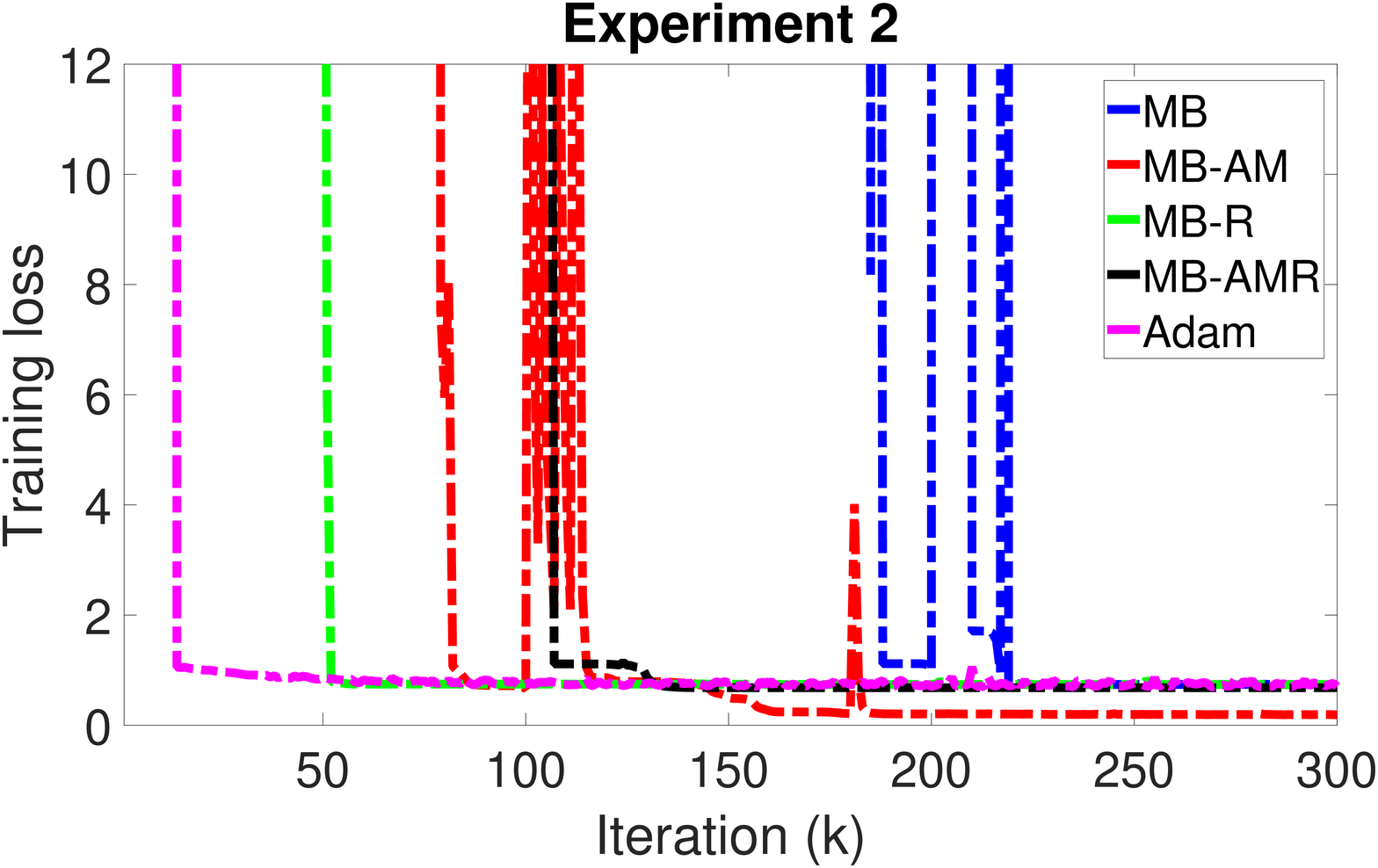} &
\includegraphics[width=0.29\textwidth]{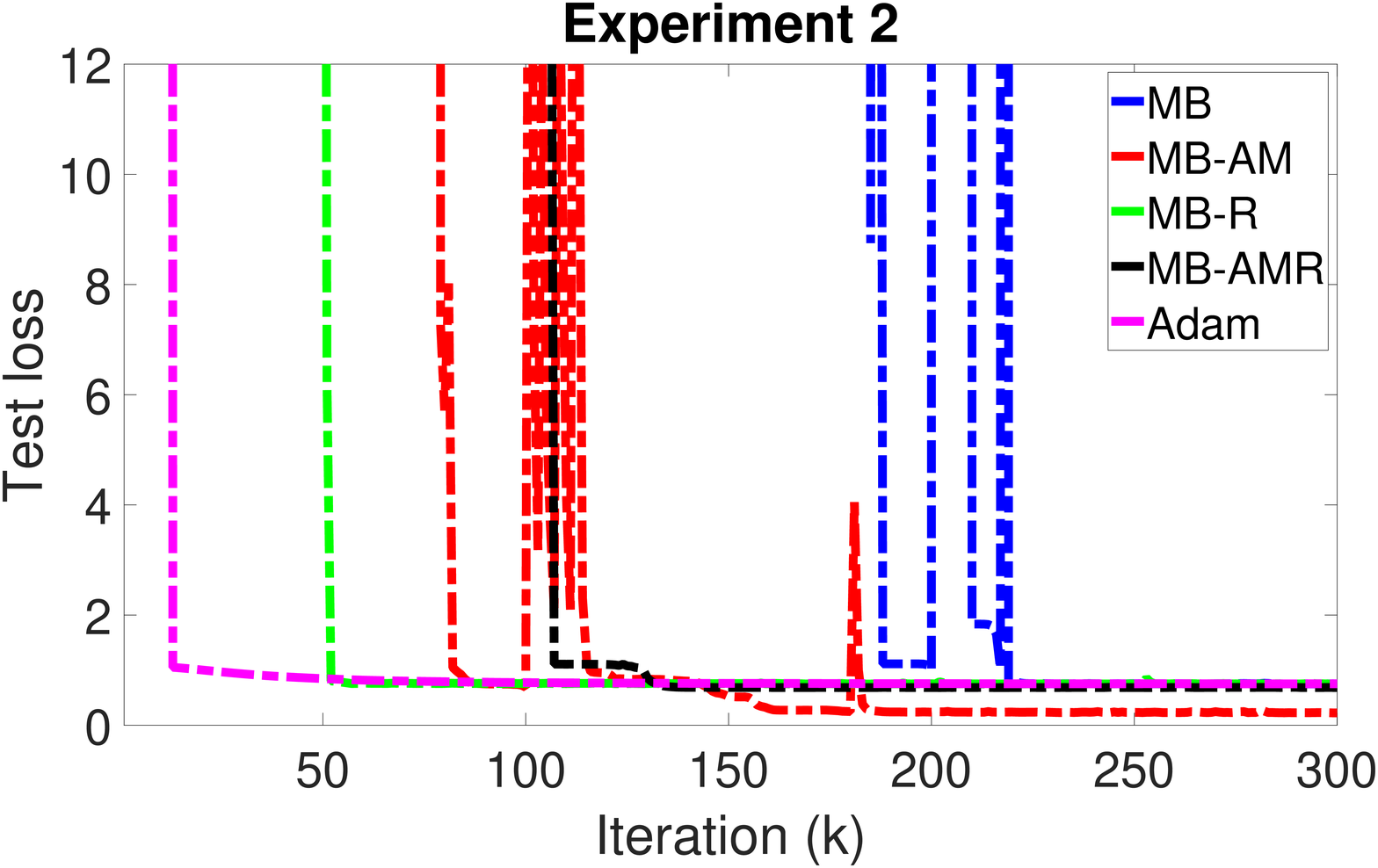} &
\includegraphics[width=0.29\textwidth]{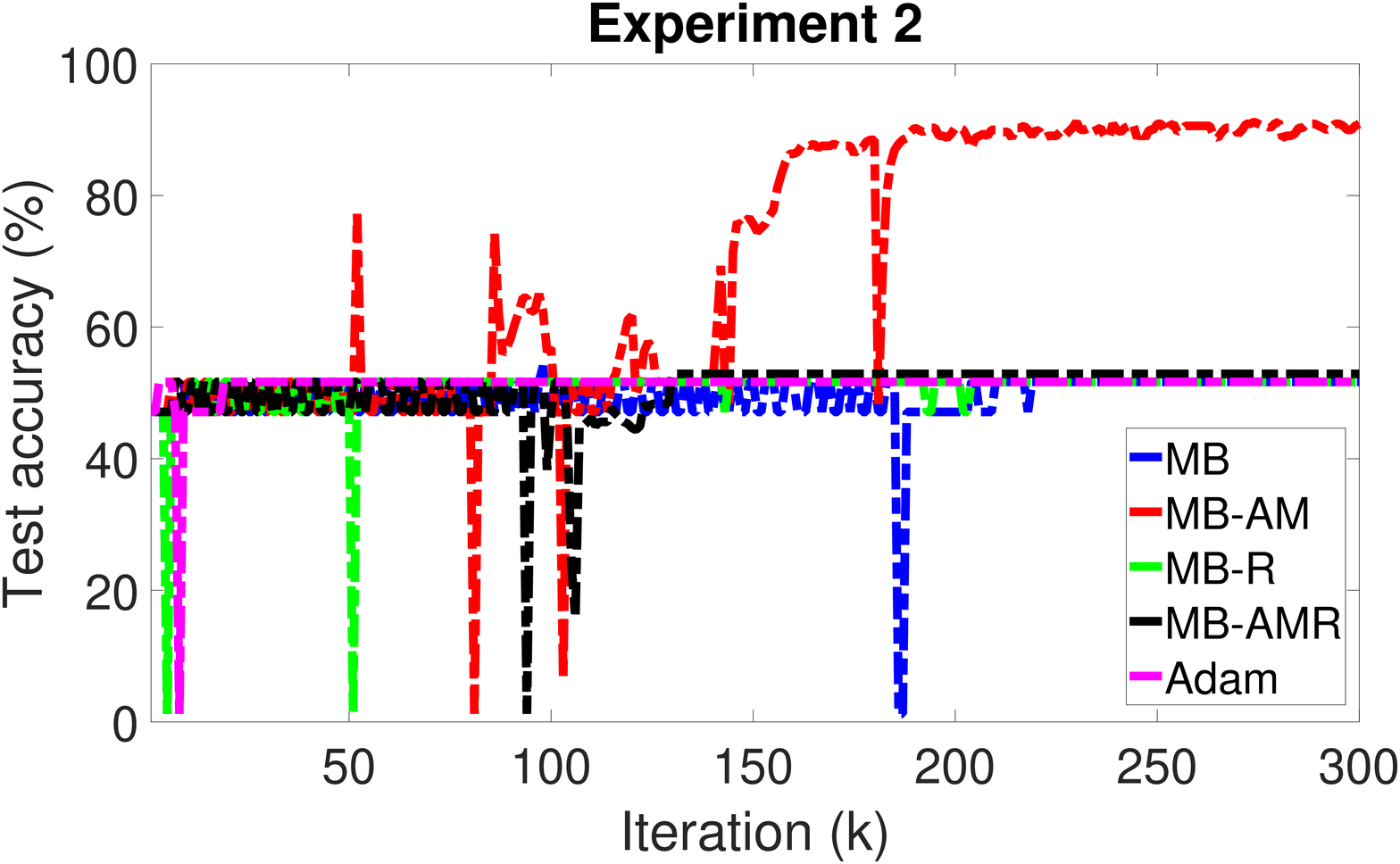} \\
\end{tabular}
\begin{tabular}{ccc}
\includegraphics[width=0.29\textwidth]{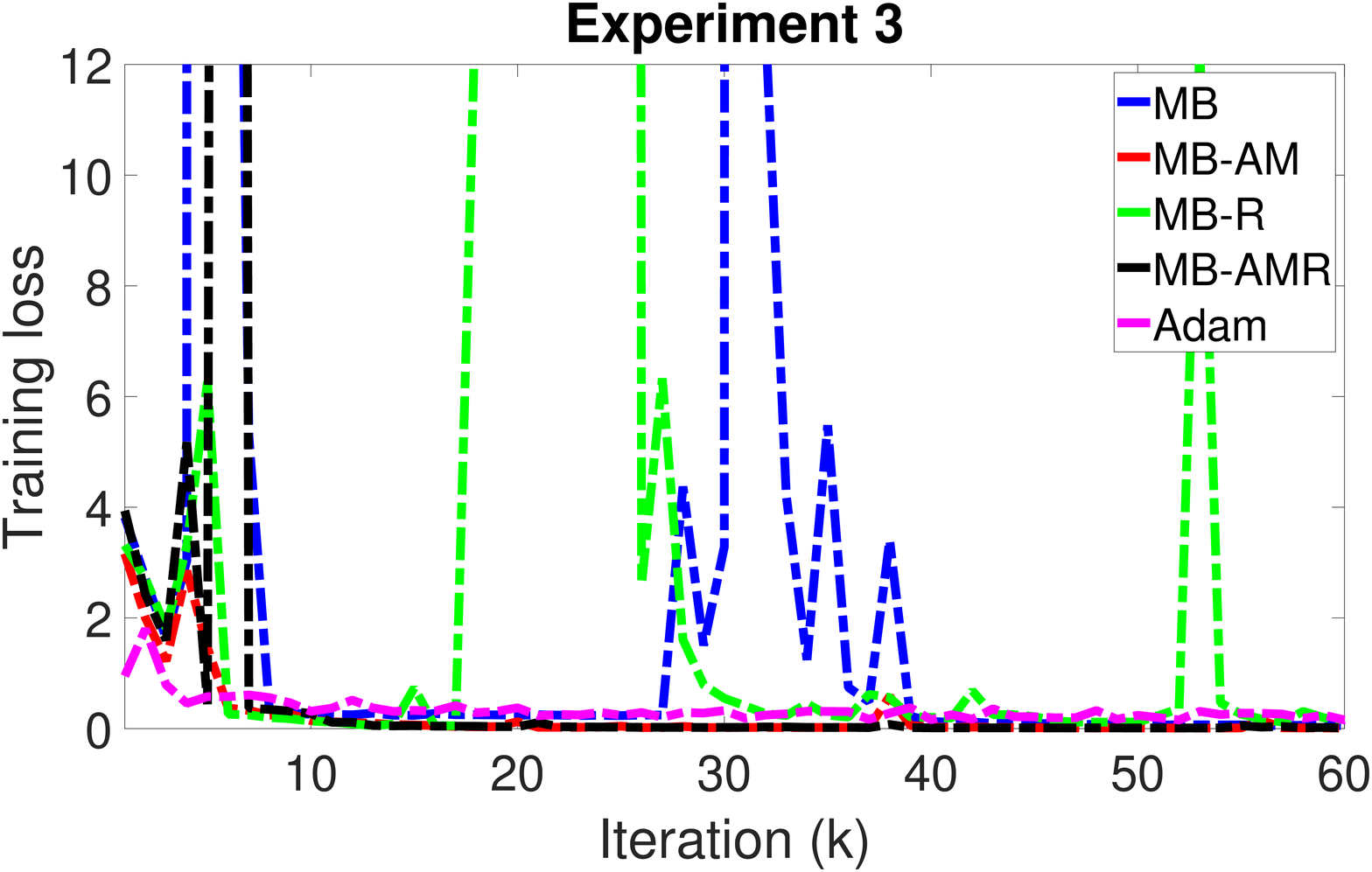} &
\includegraphics[width=0.29\textwidth]{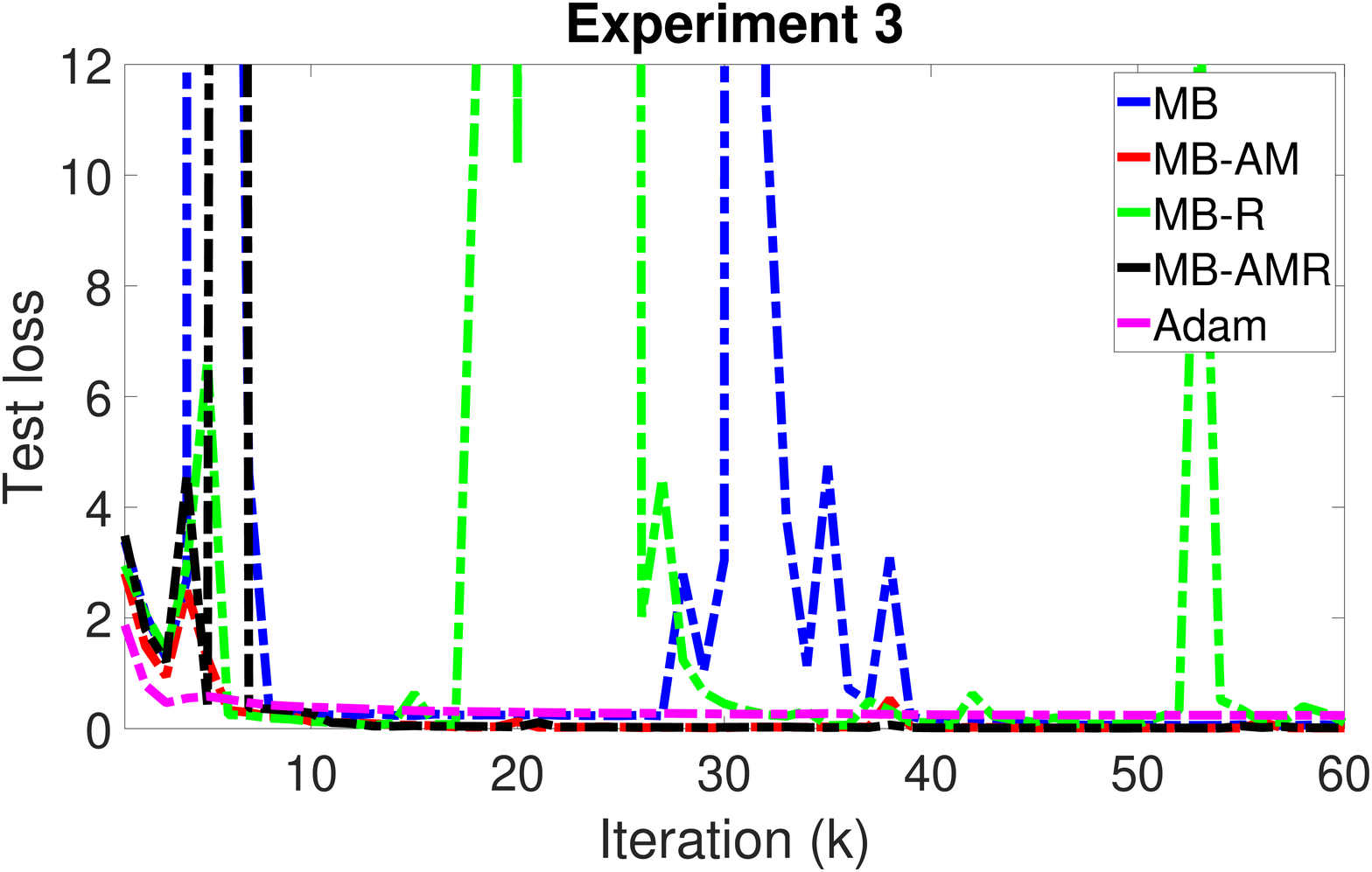} &
\includegraphics[width=0.29\textwidth]{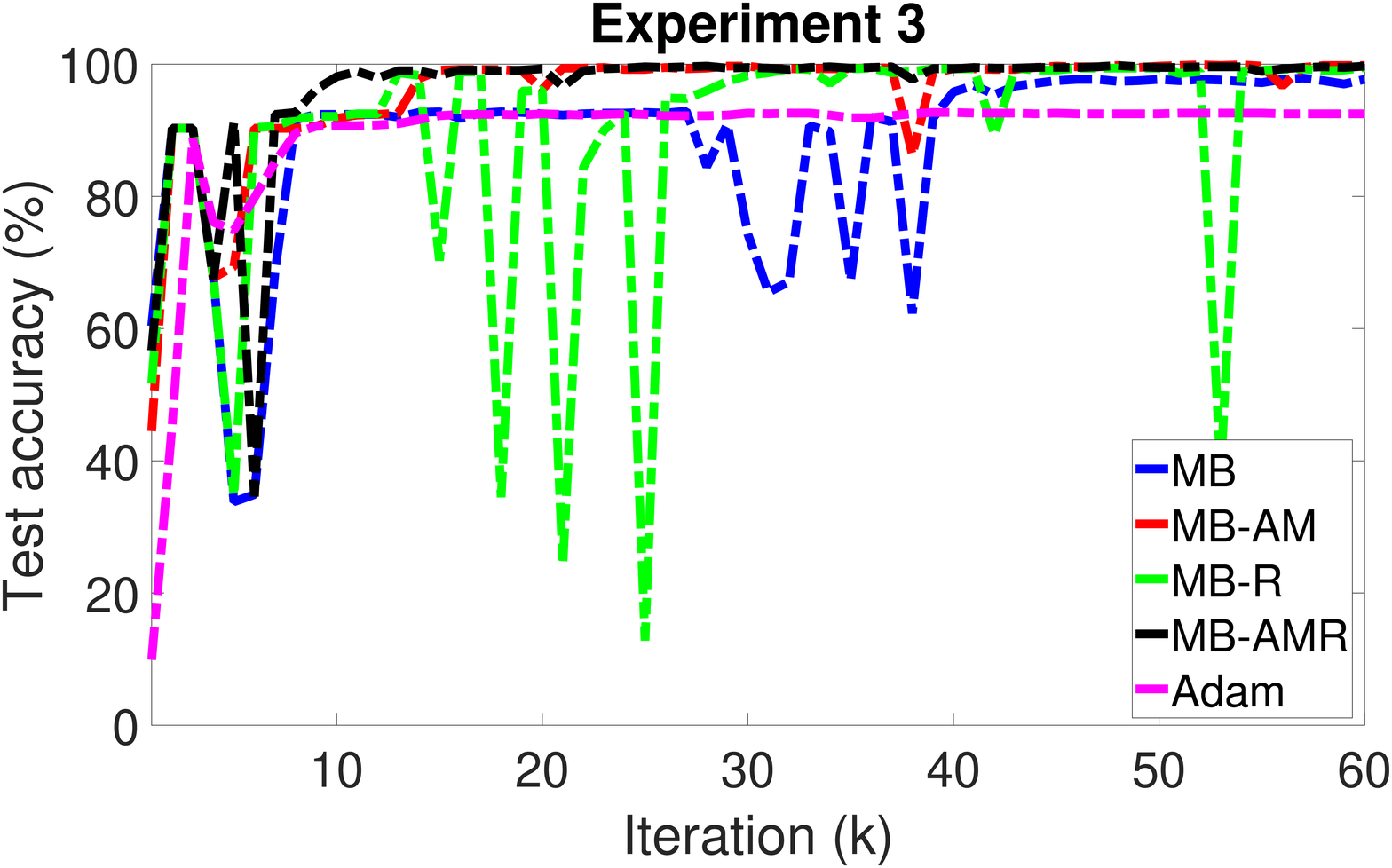} \\
\end{tabular}
\begin{tabular}{ccc}
\includegraphics[width=0.29\textwidth]{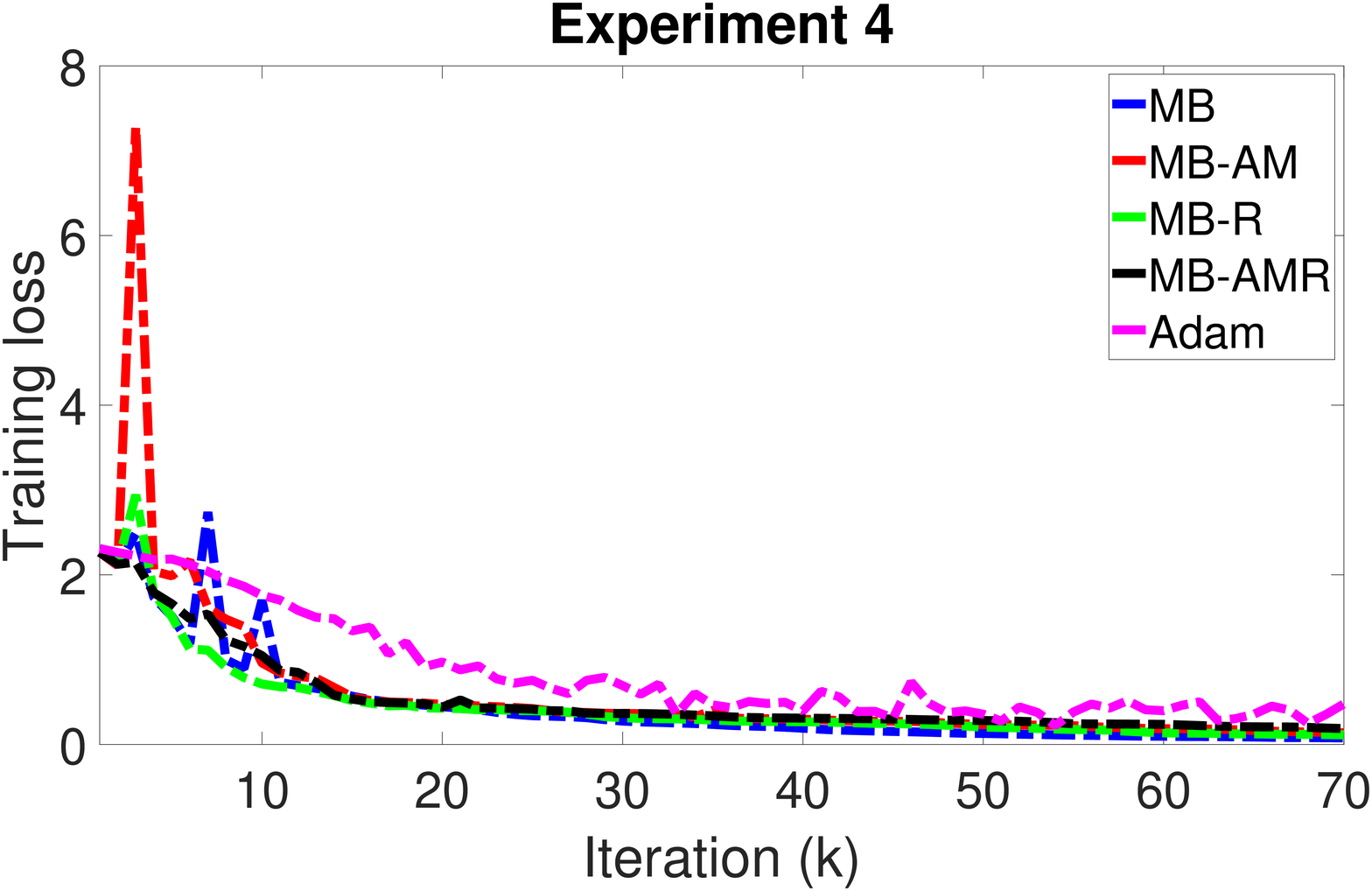} &
\includegraphics[width=0.29\textwidth]{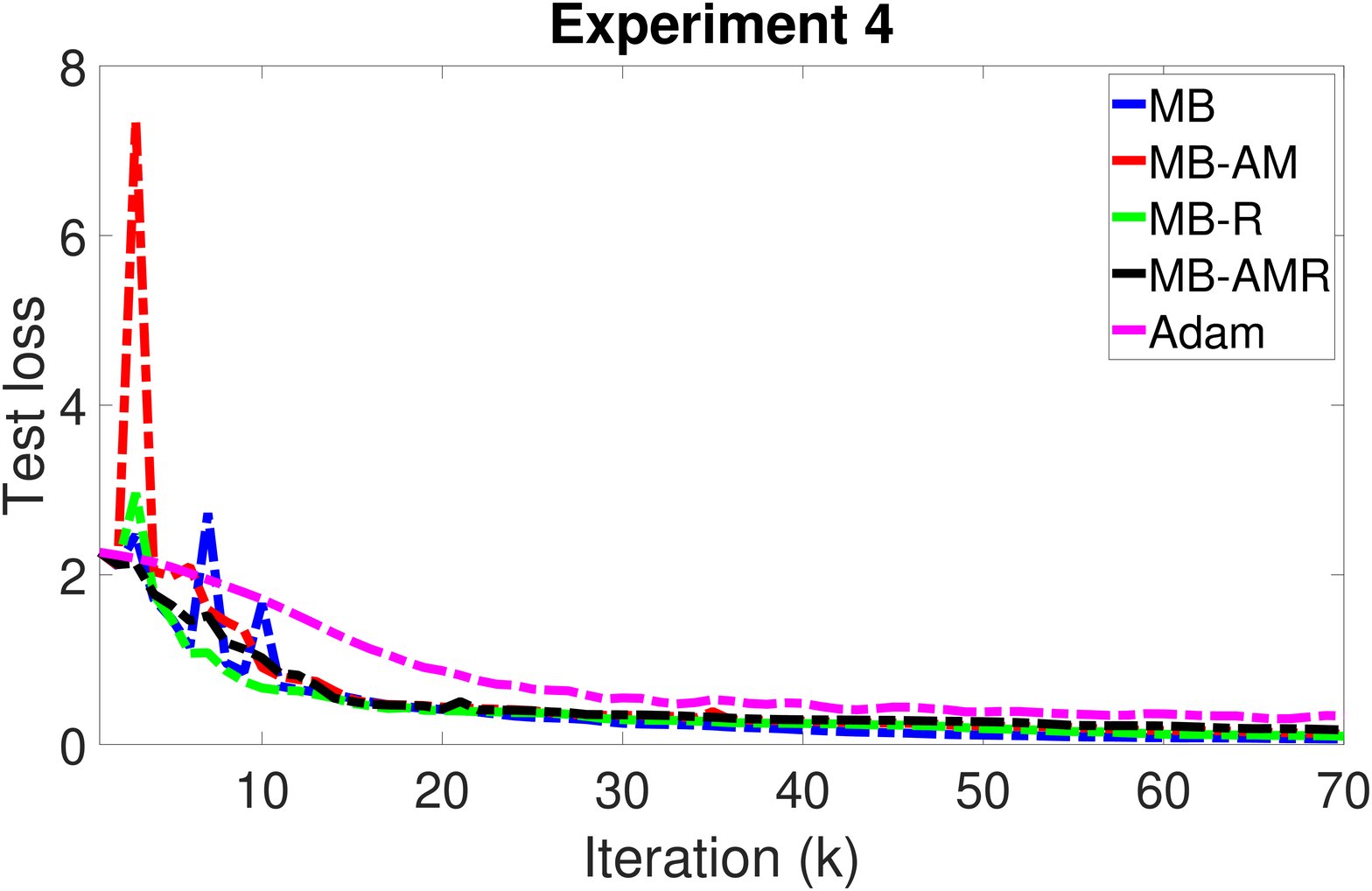} &
\includegraphics[width=0.29\textwidth]{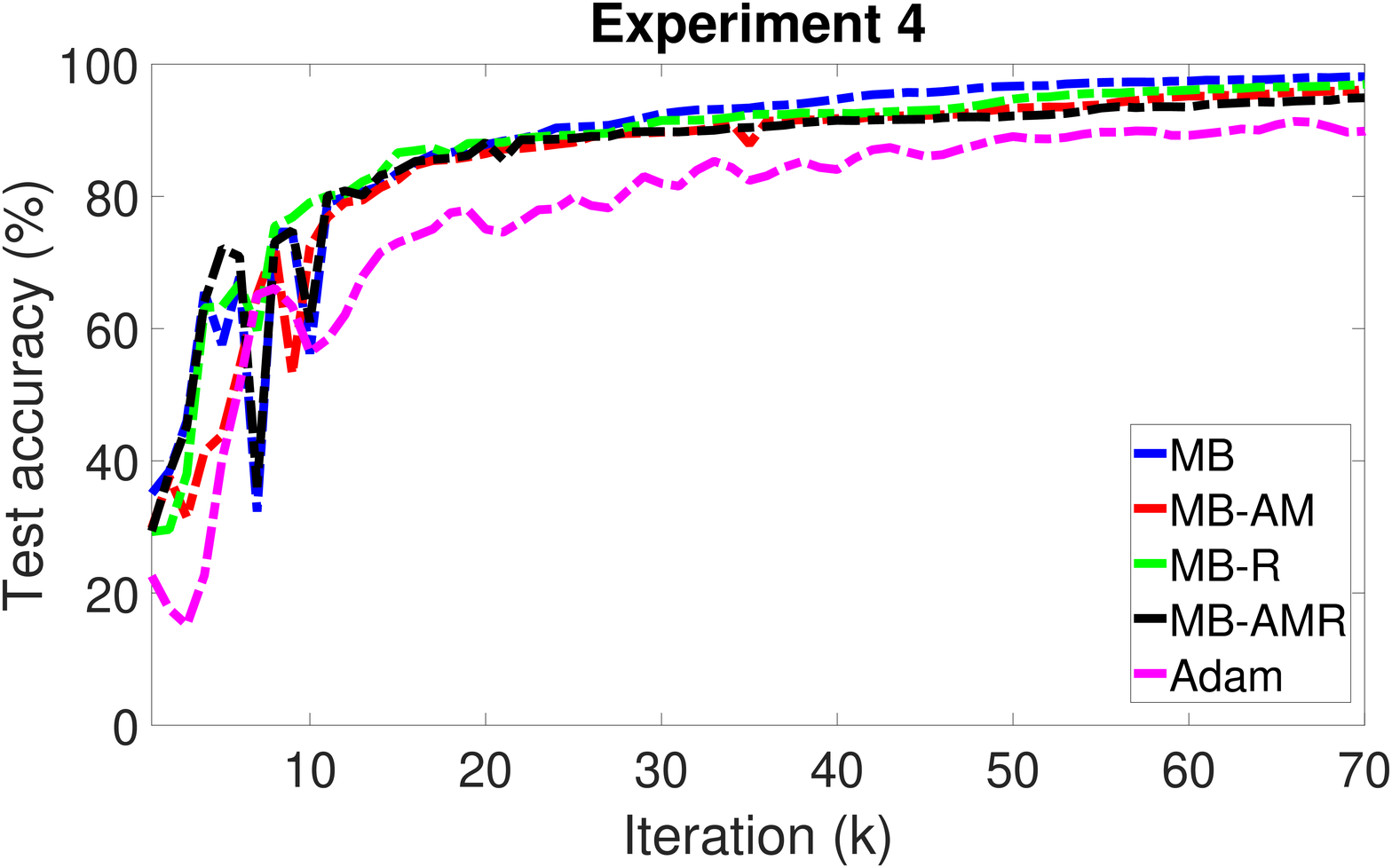} \\
\end{tabular}
\begin{tabular}{ccc}
\includegraphics[width=0.29\textwidth]{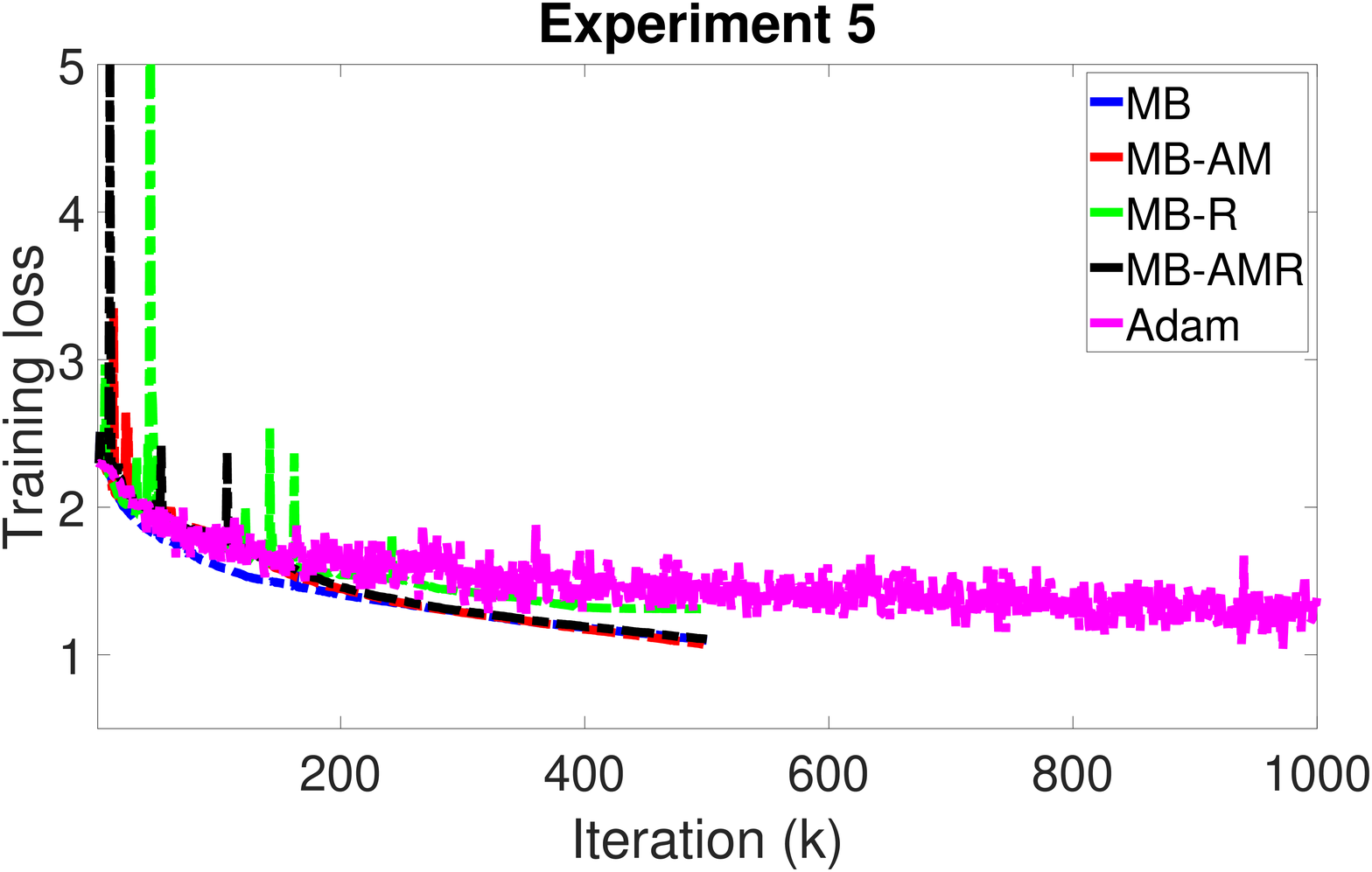} &
\includegraphics[width=0.29\textwidth]{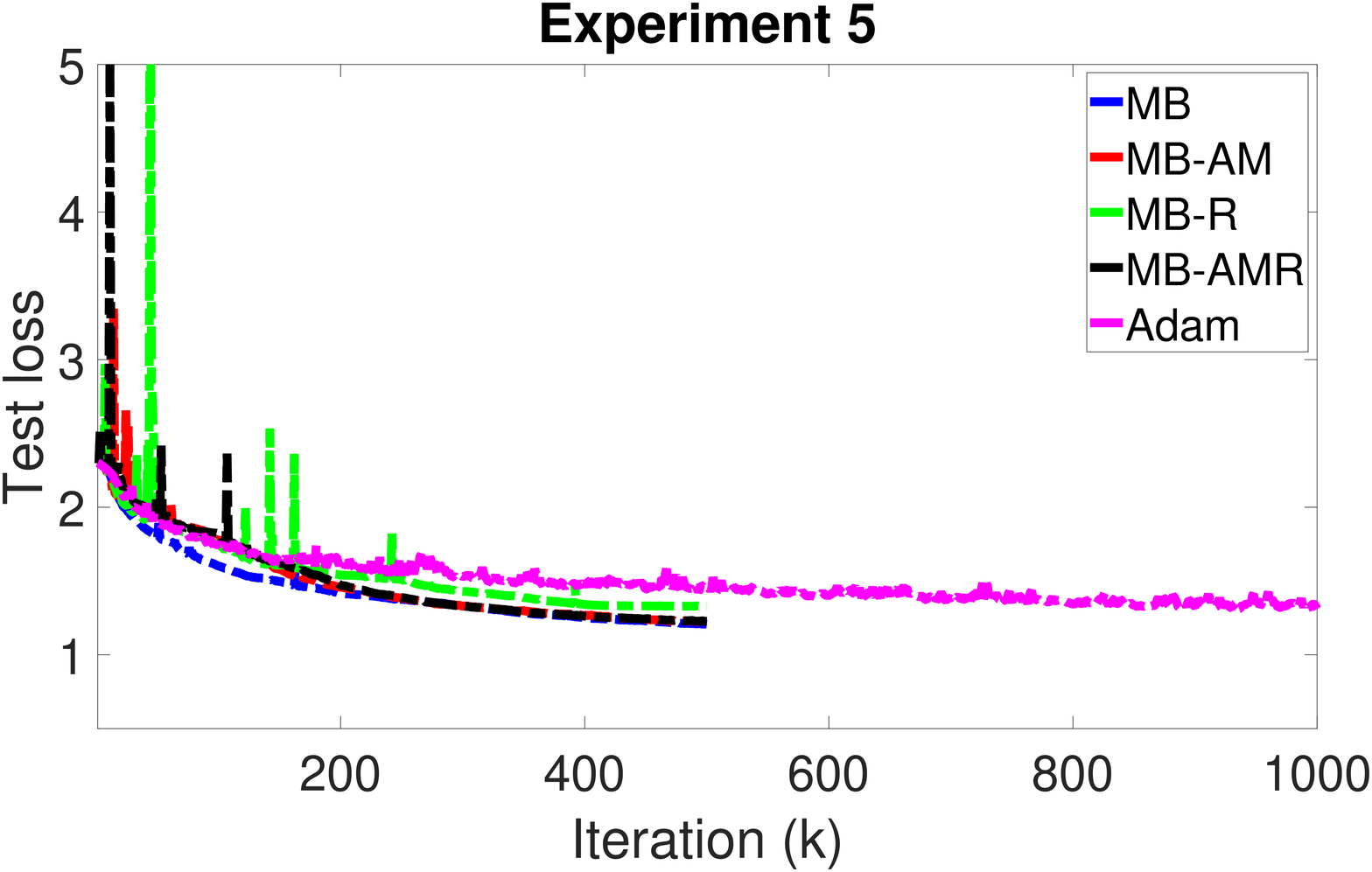} &
\includegraphics[width=0.29\textwidth]{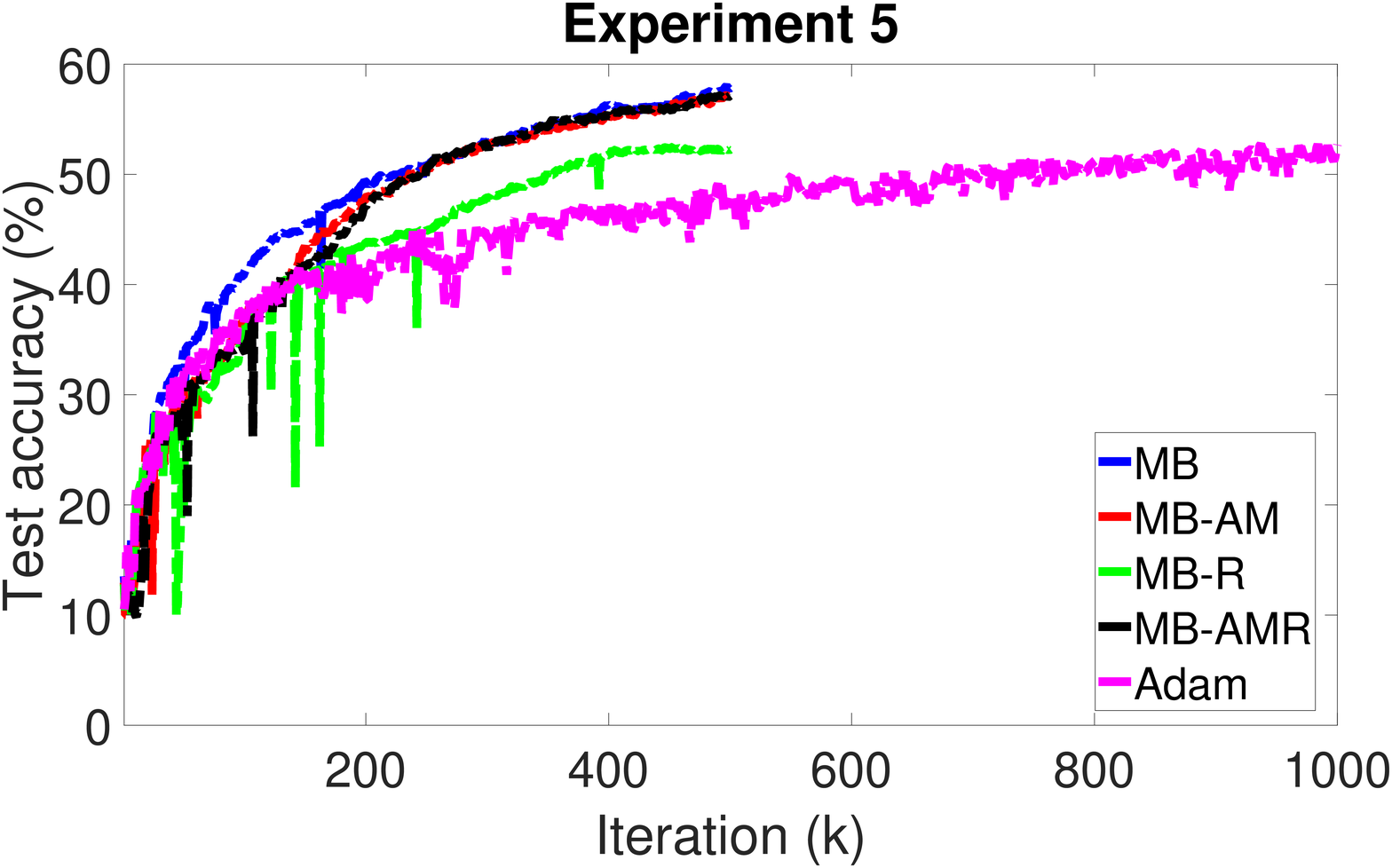} \\
\end{tabular}
\begin{tabular}{ccc}
\includegraphics[width=0.29\textwidth]{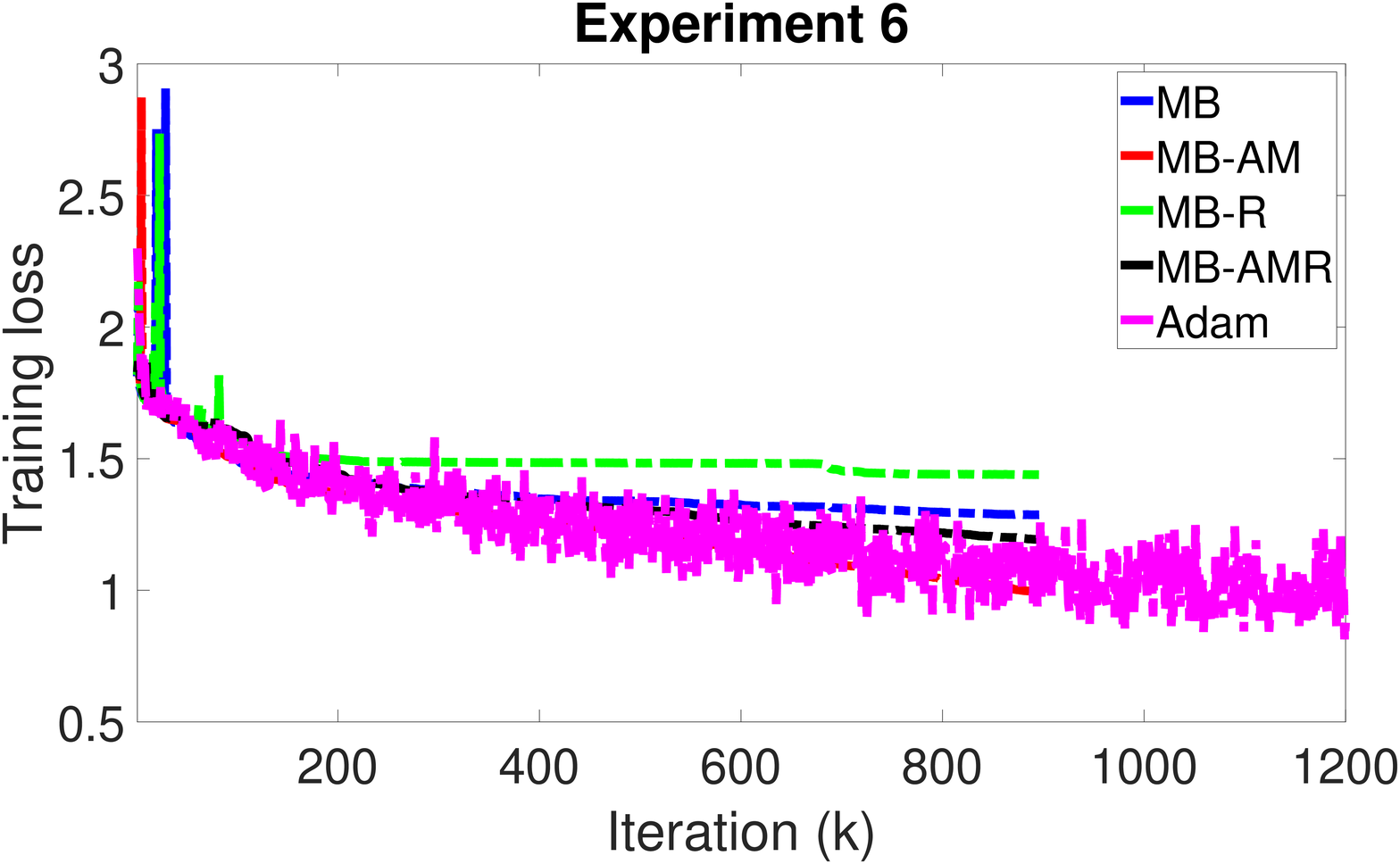} &
\includegraphics[width=0.29\textwidth]{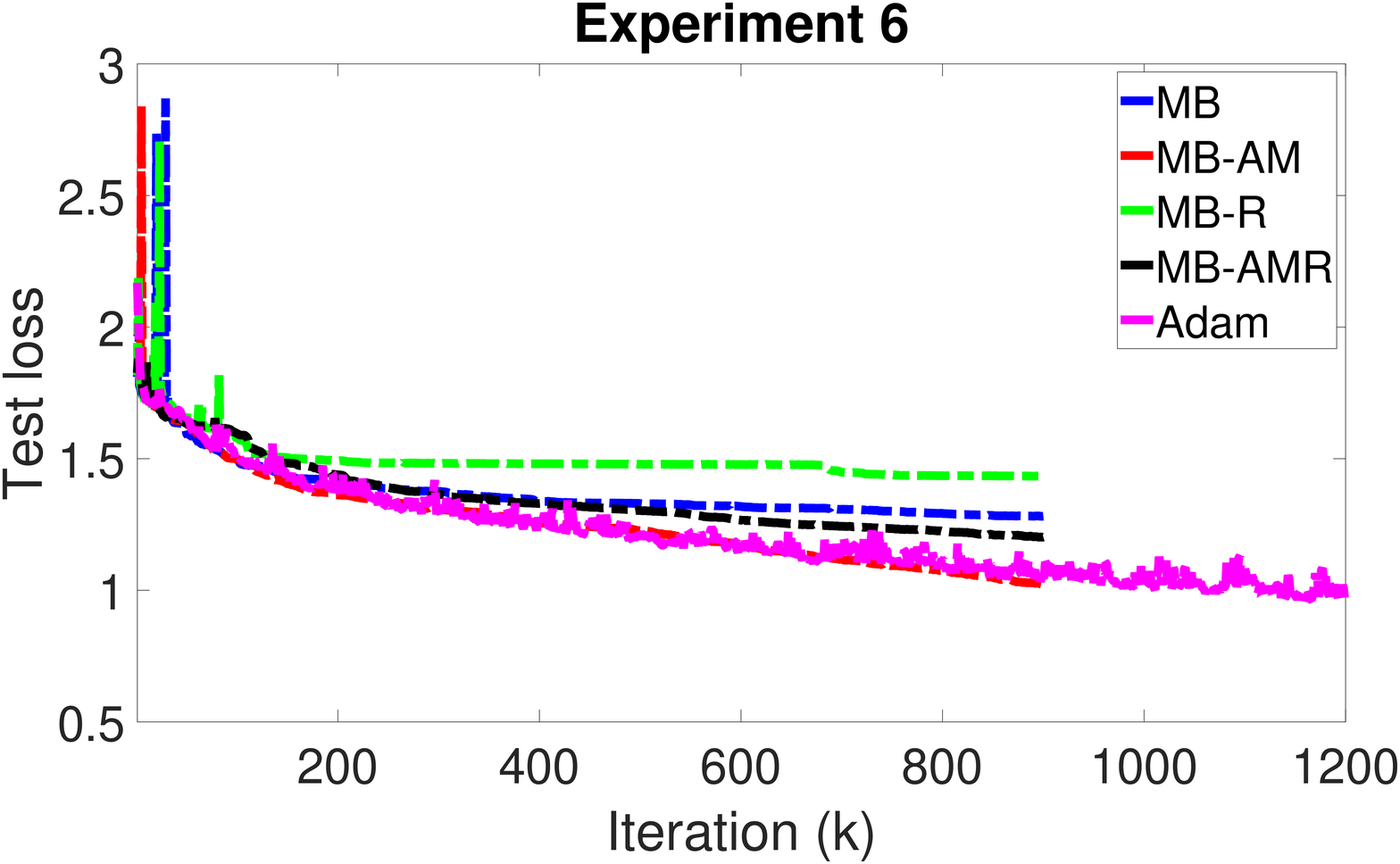} &
\includegraphics[width=0.29\textwidth]{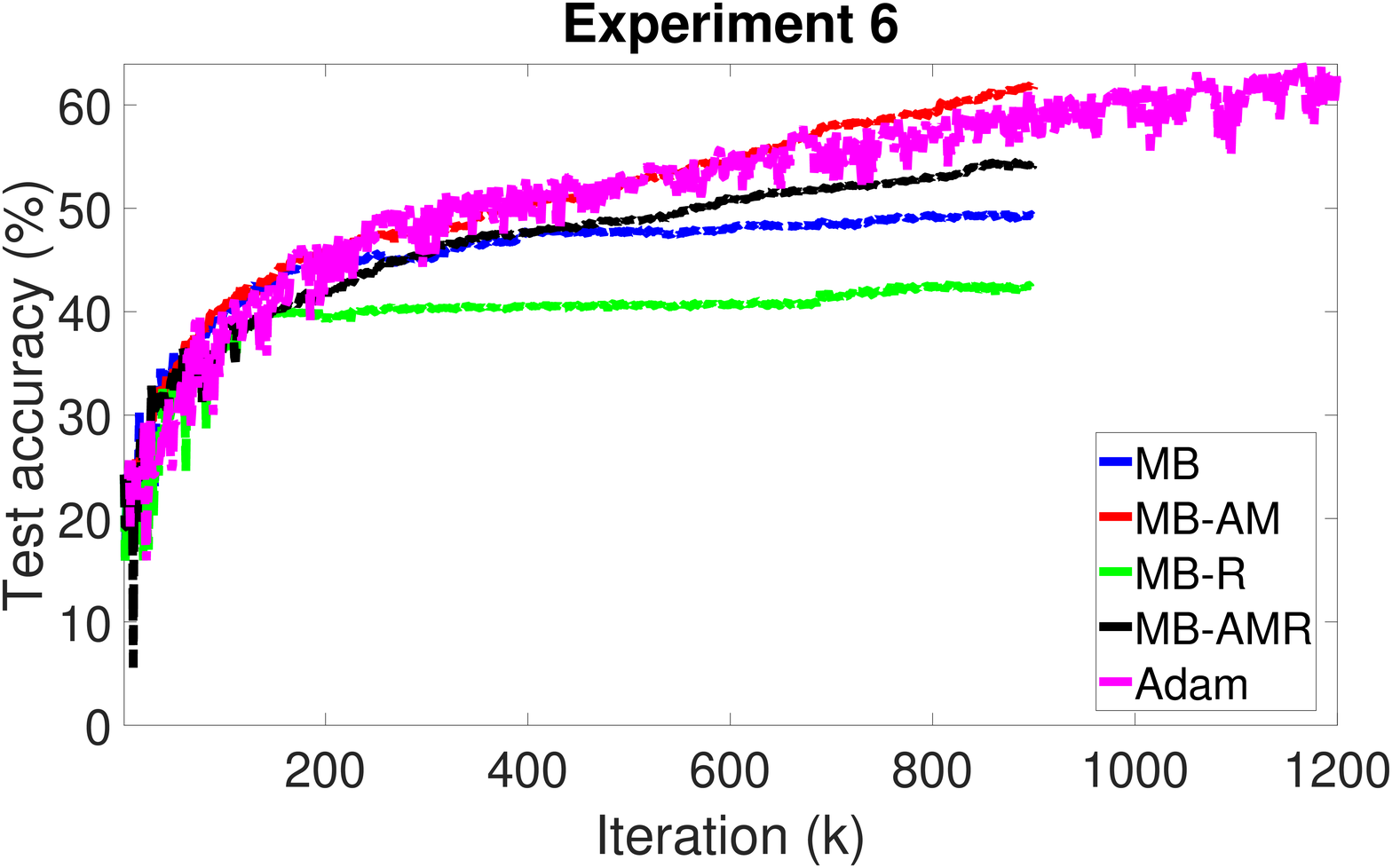} \\
\end{tabular}
\caption{The evolution of the training and test dataset cross-entropy loss and the test dataset CCR for a single simulation of each the six experiments considered in the study.}
\label{fig:6x3figure}
\end{center}
\end{figure*}

Fig. \ref{fig:memorySize} shows $m_k$ and the number of stored curvature pairs as a function of the number of iterations, $k$, while Fig. \ref{fig:6x3figure} shows the training loss, the test loss and the CCR as a function of $k$ for a single training run of each experiment. Even though these figures do not permit statistically significant conclusions to be drawn because they depict a single simulation, they are useful in revealing the general behavior of the algorithms.
 
To evaluate the regularizing property of each method 60 Monte Carlo simulations were performed. Table \ref{tab:BestOf5} reports the mean and standard deviation of the final test CCR and RNK over these simulations, where ``final" means the value once the training is completed. The last row of the table gives the mean and standard deviation over all six experiments, summarising the overall performance of each training algorithm.    
\begin{table*}
\begin{center}
\caption{Mean (standard deviation) of CCR and RNK over 60 Monte Carlo simulations. The best results for each experiment are highlighted in bold.}
\label{tab:BestOf5}
\begin{tabular}{lc@{\hskip 0.04in}cc@{\hskip 0.04in}cc@{\hskip 0.04in}cc@{\hskip 0.04in}cc@{\hskip 0.04in}c} 
&\multicolumn{2}{c}{MB} & \multicolumn{2}{c}{MB-AM} & \multicolumn{2}{c}{MB-R} & \multicolumn{2}{c}{MB-AMR} & \multicolumn{2}{c}{Adam}\\ 
& CCR & RNK & CCR & RNK & CCR & RNK & CCR & RNK & CCR & RNK \\  
\hline
Exp. 1 & 80.9(15.8) & 3.3(1.48) & 85.0(13.8) & 2.5(1.44) & 84.7(13.4) & 3.0(1.36) & 85.8(12.3) & 2.6(1.32) & \textbf{90.1}(\textbf{1.4}) & \textbf{1.9}(\textbf{1.23}) \\
\hline
Exp. 2 & \textbf{65.0}(\textbf{21.3}) & \textbf{2.3}(\textbf{1.45}) & 61.5(15.5) & 2.6(1.14) & 59.6(15.7) & 2.4(1.17) & 58.2(19.8) & 2.7(1.31) & 53.5(0.0) & 4.1(0.87) \\ 
\hline
Exp. 3 & 87.5(21.6) & 3.8(1.22) & 91.5(19.3) & 2.7(1.46) & 95.9(9.8) & 2.9(1.40) & 95.0(12.5) & \textbf{2.2}(\textbf{1.43}) & \textbf{97.8}(\textbf{1.1}) & 3.3(1.07) \\ 
\hline
Exp. 4 & 87.6(26.1) & \textbf{1.8}(\textbf{1.24}) & \textbf{92.4}(\textbf{15.4}) & 2.8(1.14) & 85.0(28.2) & 2.7(1.23) & 92.1(15.2) & 3.1(1.05) & 91.6(0.9) & 4.6(0.72) \\ 
\hline
Exp. 5 & 57.9(9.2) & \textbf{1.9}(\textbf{1.10}) & \textbf{59.5}(\textbf{1.8}) & 2.0(0.75) & 48.9(9.3) & 4.6(0.52) & 57.8(6.8) & 2.3(1.04) & 53.1(1.2) & 4.1(0.64) \\ 
\hline
Exp. 6 & 49.6(12.2) & 3.4(1.03) & 55.7(5.6) & 2.7(0.89) & 39.9(5.0) & 4.9(0.37) & 55.5(6.4) & 2.6(0.89) & \textbf{62.4}(\textbf{1.6}) & \textbf{1.3}(\textbf{0.70}) \\ 
\hline
Mean & 71.4(17.7) & 2.8(1.25) & 74.3(11.9) & \textbf{2.6}(\textbf{1.14}) & 69.0(13.6) & 3.4(1.01) & 74.1(12.2) & \textbf{2.6}(\textbf{1.17}) & \textbf{74.7}(\textbf{1.0}) & 3.2(0.87) \\
\hline
\end{tabular}
\end{center}
\end{table*}

The results vary considerably across the six experiments, with no single algorithm dominating. Adam achieves the best mean CCR performance in experiments 1, 3 and 6, MB-AM in experiments 4 and 5, and MB in experiment 2.  Overall, MB-AM, MB-AMR and Adam yield similar mean CCR performance (74.1\%-74.7\%), with Adam having much lower variance in performance across the Monte Carlo simulations than the L-BFGS methods. MB-R is the worst performing training algorithm with a CCR of 69\%, followed by MB at 71.4\%.  The fact that MB is more prone to converging to inferior solutions than the adaptive memory L-BFGS implementations is also reflected in the higher standard deviation in CCR (17.7\% versus 11.9\% for MB-AM, for example).  

In terms of the rank metric, RNK, which considers the relative performance of each algorithm and is therefore less susceptible to outliers than CCR,  MB-AM is the most consistently performing algorithm with a mean rank of 2.55, followed by MB-AMR with a mean rank of 2.58. MB is third at 2.8 and Adam is fourth at 3.2. 

Table \ref{tab:compTimes} shows the mean training computation time and, in parentheses, the mean time per iteration of the L-BFGS algorithms over the 60 Monte Carlo simulations normalized by the corresponding mean times for Adam. The algorithms have similar computation times for the three smaller dimension (MLP based) experiments, whereas MB-AM and MB-AMR are marginally faster than MB when training the larger CNN based problems (3\%-4\%). This is a consequence of the low memory usage in early iterations which speeds up the L-BFGS two-loop recursion computation. The use of large batches and the computation of the approximated second order information make the L-BFGS methods substantially more computationally intensive than Adam and the advantage of the latter increases for the larger experiments. For example, Adam is more than 22 times faster than the L-BFGS algorithms for the MNIST case study (Experiment 4). 
\begin{table}
\begin{center}
\caption{Mean training times (and in parentheses, the mean time per iteration) for each L-BFGS algorithm normalized with respect to the corresponding values for Adam.}
\label{tab:compTimes}
\begin{tabular}{c@{\hskip 0.08in}c@{\hskip 0.08in}c@{\hskip 0.08in}c@{\hskip 0.08in}c@{\hskip 0.08in}c} 
 & MB & MB-AM & MB-R & MB-AMR \\  
\hline
Exp. 1 & 2.0(2.0) & 2.1(2.1) & \textbf{1.9}(1.9) & 2.1(2.1)\\    
\hline
Exp. 2 & 5.3(5.3) & 5.8(5.8) & \textbf{5.1}(5.1) & 5.8(5.8)\\ 
\hline
Exp. 3 & 2.7(3.1) & \textbf{2.6}(\textbf{3.0}) & 2.7(3.1) & \textbf{2.6}(\textbf{3.0})\\ 
\hline
Exp. 4 & 23.2(26.5) & 22.3(25.5) & 23.0(26.2) & \textbf{22.2}(\textbf{25.4})\\ 
\hline
Exp. 5 & 15.6(31.2) & \textbf{14.8}(\textbf{29.7}) & 15.6(31.2) & 14.9(29.8)\\ 
\hline
Exp. 6 & 22.0(29.4) & \textbf{21.2}(\textbf{28.3}) & 22.0(29.3) & 21.3(28.4)\\ 
\hline
&&&& \\
\end{tabular}
\end{center}
\end{table}

\section{Conclusions}\label{sec:Concl}
This paper has proposed three variants of the multi-batch L-BFGS algorithm of \cite{berahas2016multi} based on the idea of progressive use of curvature information through a \emph{dev-increase} scheme, and periodic memory resetting as introduced in \cite{mcloone1999variable}.  The experimental results show that the \emph{dev-increase} adaptive memory scheme improves the generalisation performance and consistency of results obtained with L-BFGS  and also slightly reduces its computational complexity. Periodic memory resetting is shown to be detrimental to performance, with MB-AMR marginally inferior to MB-AM and MB-R substantially inferior to MB. The overall ordering of algorithms in terms of generalisation performance is MB-AM, MB-AMR, MB,  Adam and  MB-R. In contrast, in terms of computational efficiency the order is Adam, MB-AM, MB-AMR, MB-R, and MB, with Adam vastly superior to the other training algorithms.  Therefore, Adam remains the algorithm of choice for training large neural networks, and only when the resulting accuracy is not satisfactory, should the use of MB-AM be considered.  

MB-AM has the drawback of requiring three additional hyperparameters compared to MB ($\alpha$, $m_0/m_{max}$ and $m_{val}$). The values used in this study are empirically defined and appropriate for a first investigation of the method.  Future research will look at the automated selection of these parameters, and the potential for integrating MB-AM with progressive batching (\cite{bollapragada2018progressive}).

\bibliography{ifacconf}             
\end{document}